\documentclass[AMA,Times1COL,hidelinks]{WileyNJDv5} 

\usepackage{amsmath}
\usepackage{amssymb}
\usepackage{amsfonts}
\usepackage{amsthm}
\usepackage{galois}
\usepackage{stmaryrd}
\usepackage{float}
\usepackage{algorithm}
\usepackage{algpseudocode}
\usepackage[all,pdf]{xy}
\usepackage{lmodern,amssymb}
\usepackage[makeroom]{cancel}
\usepackage{arydshln}
\usepackage{tikz}
\usepackage{multirow}
\usetikzlibrary{shapes}
\usetikzlibrary{backgrounds}
\usepackage{caption,subcaption}

\usepackage{pifont}
\newcommand{\cmark}{\ding{51}}
\newcommand{\xmark}{\ding{55}}
\newcommand{\qmark}{\textbf{?}}

\newcommand{\D}{\mathcal{D}}

\newcommand{\R}{\mathbb{R}}
\newcommand{\I}{\mathcal{I}}

\newcommand{\FP}{\mathcal{FP}}
\newcommand{\PFP}{\powerset{\FP}}
\newcommand{\PD}{\powerset{\D}}

\newcommand{\abs}[1]{\widehat{#1}}
\newcommand{\absvec}[1]{\widehat{\uppercase{#1}}}
\newcommand{\A}{\mathcal{A}}

\newcommand{\powerset}[1]{\raisebox{.15\baselineskip}{\Large\ensuremath{\wp}}(#1)}

\DeclareMathOperator*{\argmax}{arg\,max}
\DeclareMathOperator*{\argmin}{arg\,min}

\newcommand*{\Not}{\textbf{not }}
\newcommand*\Let[2]{\State #1 $\gets$ #2}

\algblockdefx[ForEach]{ForEach}{EndForEach}[1]{\textbf{for each} #1 \textbf{do}}{\textbf{end for}}

\newif\ifboldnumber
\newcommand{\boldnext}{\global\boldnumbertrue}

\algrenewcommand\alglinenumber[1]{%
  \footnotesize\ifboldnumber\colorbox{lightgray}{#1}\else#1\fi\global\boldnumberfalse:}

\articletype{}%

\begin{document}

\title{Finding Minimum-Cost Explanations for Predictions made by Tree Ensembles}

\author[1,2]{John Törnblom}
\author[1]{Emil Karlsson}

\author[2]{Simin Nadjm-Tehrani}

\authormark{Törnblom \textsc{et al.}}
\titlemark{Finding Minimum-Cost Explanations for Predictions made by Tree Ensembles}

\address[1]{\orgname{Saab AB}, \orgaddress{Bröderna Ugglas gata
\state{Linköping}, \country{Sweden}}}

\address[2]{\orgdiv{Dept. of Computer and Information Science}, 
\orgname{Linköping University}, \orgaddress{\state{Linköping}, 
\country{Sweden}}}

\corres{John Törnblom. \email{john.tornblom@liu.se}}

\fundingInfo{
Wallenberg AI, Autonomous 
Systems and Software Program (WASP) funded by the Knut and Alice Wallenberg
Foundation.
}

\abstract[Abstract]{
The ability to reliably explain why a machine learning model arrives at a
particular prediction is crucial when used as decision support by human
operators of critical systems. The provided explanations must be provably
correct, and preferably without redundant information, called minimal
explanations. In this paper, we aim at finding explanations for predictions
made by tree ensembles that are not only minimal, but also minimum with
respect to a cost function. 
To this end, we first present a highly efficient oracle that can determine the 
correctness of explanations, surpassing the runtime performance of current 
state-of-the-art alternatives by several orders of magnitude. 
Secondly, we adapt an algorithm called MARCO from the optimization field 
(calling the adaptation m-MARCO) to compute a single minimum explanation per
prediction, and demonstrate an overall speedup factor of 2.7 compared to a
state-of-the-art algorithm based on minimum hitting sets, and a speedup factor
of 27 compared to the standard MARCO algorithm which enumerates all minimal 
explanations.

Finally, we study the obtained explanations from a range of use cases, leading
to further insights into their characteristics. In particular, we observe that
in several cases, there are more than 500,000 minimal explanations to choose
from for a single prediction. In these cases, we see that only a small portion
of the minimal explanations are also minimum, and that the minimum explanations
are significantly less verbose, hence motivating the aim of this work. 
}


\jnlcitation{\cname{%
\author{Törnblom J.},
\author{Karlsson E}, and
\author{Nadjm-Tehrani S}}.
\ctitle{Finding minimum-cost explanations for predictions made by tree ensembles.}
\cjournal{\it Softw: Pract Exper.} \cvol{2025;00(00):1--18}.
}

\maketitle

\section{Introduction}
\label{sec:introduction}
In many practical applications of machine learning, the ability to explain why
a prediction model arrives at a certain decision is crucial\cite{Hadji21}.
Although significant progress has been made to understand the intricacies of 
machine learning systems and how their predictions can be 
explained\cite{Ras22,Kaur22}, 
there are still open research questions. In this work, we are concerned with the 
use of machine learning models in critical decision support systems such as 
medical diagnosis and flight management systems, where humans act in a loop,
working together with the system to make critical decisions. In such cases,
explanations for predictions enable their system operators to make informed 
interventions of autonomous actions when explanations appear unjustified. More
specifically, we aim to provide explanations that are provably correct and 
without redundant information for predictions made by a class of machine 
learning models called tree ensembles.
To illustrate the notion and usefulness of such explanations, consider a fictive
bank-loan application support system, realized by the simple decision tree 
depicted in Figure~\ref{fig:bankloan}.

\begin{figure}[ht]
  \centering
\begin{tikzpicture}[nodes={ellipse,draw}, ->, scale=1]
    \tikzstyle{every node}=[draw=black, ellipse, align=center, thin]
    \tikzstyle{level 1}=[level distance=13mm, sibling distance=53mm]
    \tikzstyle{level 2}=[sibling distance=33mm] 
    \node {low income}
        child {node {criminal record}
            child {node {low education}
                child {node[rectangle] {0}
                    edge from parent node[left,draw=none] {no}
                }
                child {node {missed payments}
                    child {node[rectangle] {0}
                        edge from parent node[left,draw=none] {no}
                    }
                    child {node[rectangle] {1}
                        edge from parent node[right,draw=none] {yes}
                    }
                    edge from parent node[right,draw=none] {yes}
                }
                edge from parent node[left,draw=none] {no}
            }
            child {node[rectangle] {1}
                edge from parent node[right,draw=none] {yes}
            }
            edge from parent node[left,draw=none] {no}
        }
        child {node {low education}
            edge from parent 
            child {node[rectangle] {0}
                edge from parent node[right,draw=none] {no}
            }
            child {node {missed payments}
                child {node[rectangle] {0}
                    edge from parent node[left,draw=none] {no}
                }
                child {node[rectangle] {1}
                    edge from parent node[right,draw=none] {yes}
                }
                edge from parent node[right,draw=none] {yes}
            }
            edge from parent node[right,draw=none] {yes}
        }
        ;
  \end{tikzpicture}
  \caption{A fictive bank-loan application processing system, where a 
           positive outcome indicates that the applicant should be denied a
           loan.}
  \label{fig:bankloan}
\end{figure}
Suppose the system processes a particular applicant with a low income, low 
education, no criminal records, and a couple of missed payments, and that the 
system denies the applicant a bank loan. When explaining to the applicant 
why the requested loan is denied, information about criminal records may be
omitted since that particular feature does not contribute to the outcome of 
the prediction. Furthermore, if the applicant requests a new assessment after 
getting a salary raise (so that the applicant no longer has a low income), the 
system would respond the same. Consequently, an explanation only needs to 
include information about two of the four features: low education and missed
payments. We say that explanations that are correct and without redundant
information are minimal. Typically, there are several minimal explanations to
choose from, potentially hundreds of thousands for large systems. By 
associating each minimal explanation with a cost computed by a domain-specific
cost function, we can order them to find one that is the most desirable for the 
problem at hand, called a minimum explanation. To be useful in decision support
systems, the produced explanation should also be calculated reasonably fast.
To this end, this paper contributes with the following.

\begin{itemize}
\item We formalize a sound and complete oracle for determining the 
  correctness of an explanation in the abstract interpretation 
  framework, designed specifically for tree ensembles.
          
\item We demonstrate the performance of our oracle in a case study from
  related work that computes minimal explanations, and conclude an 
  overall speedup factor of 2,500 compared to the current state-of-the-art.
          
\item We demonstrate that, in the presence of a highly efficient oracle, 
  it is possible to enumerate all minimal explanations for several 
  predictions made by non-trivial tree ensembles.
  
\item We propose an algorithm called m-MARCO which is an adaptation of the
  standard MARCO algorithm the field of optimization for the purpose of
  computing an explanation that is minimum, and demonstrate an overall
  speedup factor of 27 compared to the MARCO algorithm which enumerates
  all minimal explanations.

\item When paired with our efficient oracle, we demonstrate that m-MARCO
  is significantly faster than two other competing algorithms: one based
  on a branch-and-bound approach (BB), and another one based on minimum 
  hitting sets (MHS).
\end{itemize}
We also provide\footnote{https://github.com/john-tornblom/VoTE-explain-experiments} implementations of our
algorithms, together with logs and automated scripts to reproduce our results.

The rest of this paper is structured as follows. Section~\ref{sec:prel}
provides preliminaries on decision trees and tree ensembles, the computation 
of minimal and minimum explanations, the MARCO algorithm, and abstract 
interpretation. In Section~\ref{sec:related-works}, we relate our contributions
to earlier works. In Section~\ref{sec:oracle}, we formalize an oracle in the
abstract interpretation framework that can determine the correctness of 
explanations, and in Section~\ref{sec:algos} we formalize algorithms for 
computing minimal and minimum explanations. In Section~\ref{sec:comp-study}, 
we study the runtime performance of the formalized algorithms, and investigate 
characteristics of the explanations computed in that study, e.g., how many 
minimal and minimum explanations there are for a given prediction. Finally, 
Section~\ref{sec:conclusions} concludes the paper. For readability, 
Sections~\ref{sec:prel}--\ref{sec:algos} are only concerned with binary 
classification problems. Supplementary algorithms for reasoning about tree 
ensembles trained on multi-class classification problems is provided in 
Appendix~\ref{sec:multi-class}.

\section{Preliminaries}
\label{sec:prel}
In this section, we bring together notions from several fields in order to 
establish a common vocabulary for the rest of the paper. First, we formalize 
the prediction function for tree ensembles trained on binary classification 
problems (a formalization for multi-class classification problems is available
in Appendix~\ref{sec:multi-class}). Next, we define the notion of valid 
explanations with respect to a particular prediction, and how we can distinguish
between them based on their verbosity. We then review the MARCO algorithm
presented elsewhere, which can be used in conjunction with an oracle to 
enumerate all minimal explanations for a particular prediction. Finally, we
communicate basic ideas that are essential for sound reasoning with abstract
interpretation, concepts that are foundational when constructing the oracle 
we later use to compute desirable explanations.

\subsection{Notation}
We use lower-case letters to denote functions and scalars, and upper-case
letters to denote sets. The number of input dimensions of functions is 
typically denoted $n$. Input variables are normally named $x$, and output 
variables $y$. A function named $f$ that accepts an argument $x$ and is 
parameterized with a model $M$ is denoted $f(x; M)$. Tuples of scalars are 
annotated with a bar, e.g., $\bar{x} = (x_1, \ldots, x_n)$, while variables and
functions associated with abstract interpretation are annotated with a hat, e.g.,
$\abs{x}$. Finally, we denote the power set over a set $S$ as $\powerset{S}$.
\subsection{Decision Trees and Tree Ensembles}
In machine learning, decision trees are used as predictive models to capture 
statistical properties of a system of interest. As the name suggests, the 
prediction function of a decision tree is parameterized by a tree structure, 
as exemplified by Figure~\ref{fig:decision-tree}. 

\begin{figure}[ht]
  \begin{minipage}{0,49\textwidth}
    \centering
    \begin{tikzpicture}[nodes={ellipse,draw}, ->]
      \tikzstyle{level 1}=[sibling distance=30mm]
      \tikzstyle{level 2}=[sibling distance=20mm] 
      \tikzstyle{every node}=[draw=black, ellipse, align=center]
      \draw node{$x_1 \leq c_1$}
      child{node{$x_2 \leq c_2$}
        child{node[rectangle] {$y_1$}}
        child{node[rectangle] {$y_2$}}
      }
      child{node{$x_2 \leq c_3$}
	    child{node[rectangle] {$y_3$}}
	    child{node[rectangle] {$y_4$}}
      }
      ;
    \end{tikzpicture}
  \end{minipage}
  \begin{minipage}{0,49\textwidth}
    \centering
    \begin{tikzpicture}[scale=0.7]
      \draw[->] (0,0) -- (7,0) node[right] {$x_1$}; 
      \draw[->] (0,0) -- (0,5) node[above] {$x_2$};
      \draw (3,0) -- (3,5);
      \draw (0,3) -- (3,3);
      \draw (3,4) -- (7,4);
      
      \draw (1.5, 1.8) node[below] {$y_1$};
      \draw (1.5, 4) node[above] {$y_2$};
      \draw (5, 2.4) node[below] {$y_3$};
      \draw (5, 4.4) node[above] {$y_4$};
      
      \draw (3, -0.05) -- (3, 0.05) node[below] {$c_1$};
      \draw (-0.05, 3) -- (0.05, 3) node[left] {$c_2$};
      \draw (-0.05, 4) -- (0.05, 4) node[left] {$c_3$};
    \end{tikzpicture}
  \end{minipage}
  \caption{A decision tree (to the left) partition an input space
    into disjoint regions (to the right), and associates each region
    with an output value.}
  \label{fig:decision-tree}
\end{figure}

The tree is evaluated in a top-down manner, where decision rules associated 
with intermediate nodes determine which path to take towards the leaves, and 
ultimately which value to output.

In general, decision rules in decision trees may be non-linear and multivariate,
and leaves can be associated with tuples of scalars. In this paper, we focus 
on trees trained by XGBoost\cite{Chen16}, in which case all decision 
rules are univariate and linear, and each leaf is associated with a single 
scalar value. Consequently, we can formalize the prediction function as a 
mapping from adjacent hyperrectangles to scalar values as follows.
\begin{definition}[Decision Tree Prediction Function]
\label{def:decision-tree}
Let $X_1, \ldots, X_k$ be a partition of an input space, 
$y_1, \ldots, y_k$ scalar values from the output space, and 
$T = \{(X_1, y_1), \ldots, (X_k, y_k)\}$.
The prediction function for a decision tree may then be defined as
\begin{equation*}
  t(\bar{x}; T) = 
  \begin{cases} 
    y_1 & \text{if } \bar{x} \in X_{1}, \\
     &  \vdots\\
    y_k & \text{if } \bar{x} \in X_{k}. \\
  \end{cases}
\end{equation*}
\end{definition}

A decision tree trained on a dataset with $n$ features may contain unreferenced
variables, i.e., the tree only considers a subset of its input variables during 
prediction. This is typically the case in high-dimensional applications, and 
when the training dataset contains statistically insignificant features. 
Since such input variables never affect predictions, they can be removed from
all explanations. Consequently, we define the notion of \textit{variables
referenced by a tree}, which we later use in Section~\ref{sec:algos} to 
improve the runtime performance when computing explanations.
\begin{definition}[Variables Referenced by a Tree]
\label{def:var-ref-tree}
The variables referenced by a tree during predictions is a set of indices
$V_t \subseteq \{1, \ldots, n\}$, defined as the union of all the dimensions 
its decision rules operate in.
\end{definition}
\subsubsection{Tree Ensembles}
Decision trees are known to suffer from overfitting, i.e., the model becomes 
too specialized towards training data, and the prediction function generalizes 
poorly when confronted with previously unseen inputs. To counteract overfitting 
of decision trees, several types of tree ensembles have been proposed, e.g., 
random forests\cite{Breiman01} and gradient boosting 
machines\cite{Friedman01}. The different types of tree ensembles are 
normally distinguished by their learning algorithms, while their prediction
functions have similar structure.
\begin{definition}[Tree Ensemble Prediction Function]
\label{def:tree-ensemble}
Let $F = (T_1, \ldots, T_b)$ be a tuple of $b \in \mathbb{Z}_{+}$ decision
trees (as defined by Definition~\ref{def:decision-tree}), all sharing the same
input and output space. The prediction function for a tree ensemble may 
then be defined as
\begin{equation*}
  \label{eq:tree-ensemble}
  f(\bar{x}; F) = \sum_{T \in F} t(\bar{x}; T).
\end{equation*}
\end{definition}

Similar to a decision tree, an ensemble of trees may also only consider a 
subset of its input variables during prediction.
\begin{definition}[Variables Referenced by a Tree Ensemble]
\label{def:var-ref-ensemble}
The variables referenced by a tree ensemble during predictions is a set of indices
$V_f \subseteq \{1, \ldots, n\}$, defined as the union of all variables
referenced by each individual tree in the ensemble.
\end{definition}

\subsubsection{Classification with Tree Ensembles}
\label{sec:classification}
By training tree ensembles to predict probabilities, they can be used as 
classifiers. Leaves in classifiers trained by XGBoost are associated with 
values from the logarithmic domain that are summed together, followed by a 
transformation to the domain of probabilities using the sigmoid function.
\begin{definition}[Sigmoid Function]
\label{def:sigmoid}
The sigmoid function $p_{\sigma}$ is a monotone function that transforms a 
real-valued scalar $z$ into an output value in the range $[0, 1]$, and is
defined as
\begin{equation*}
  \label{eq:sigmoid}
  p_{\sigma}(z) = \frac{1}{1 + e^{-z}}.
\end{equation*}
\end{definition}
With the sigmoid function defined, we can define the tree-based binary 
classifier.
\begin{definition}[Tree-based Binary Classifier]
\label{def:binary-classifier}
Let $F$ be a tree ensemble trained to predict the probability that a given
input tuple $\bar{x}$ maps to one of two classes. A binary classifier $f_{bin}$
that discriminates between the two classes may then be defined as
\begin{equation*}
  \label{eq:binary-classifier}
  f_{bin}(\bar{x}; F) = p_{\sigma}\big(f(\bar{x}; F)\big) > 0.5.
\end{equation*}
\end{definition}

Tree ensembles can also be used for multi-class classifications, i.e., to
discriminate between three or more classes.  For readability, however,
Sections~\ref{sec:prel}--\ref{sec:algos} of this paper are only concerned
with binary classification problems. For the sake of completeness, a 
formalization of multi-class classifiers is defined in 
Appendix~\ref{sec:multi-class}.

\subsubsection{Running Example}
\label{sec:cls-example}
For illustrative purposes, we use a simple tree-based binary classifier trained
on the DREBIN dataset\cite{Arp14}, a dataset extracted from a large set of
Android applications labeled as either malware or benign, throughout this paper. 
The classifier was trained using XGBoost on a Boolean feature space 
that encodes permissions requested by the application, yielding a tree ensemble 
$F = \{T_1, T_2, T_3\}$ as depicted in Figure~\ref{fig:drebin-model}. 
\begin{figure}[ht]
  \centering
  \begin{minipage}{0,32\textwidth}
    \begin{tikzpicture}[->]
      \tikzstyle{level 1}=[sibling distance=20mm, level distance=20mm]
      \tikzstyle{level 2}=[sibling distance=10mm]
        \begin{scope}
        \node[scale=1.3] at (0,4) {$T_1$};
        \draw[rounded corners] (-2.6, -1.4) 
             rectangle (2.2 ,4.4)
        ;
      \end{scope}
      \begin{scope}[yshift=3cm]
      \draw 
        node[draw,ellipse,align=center,scale=0.8]{send SMS\\$(x_1)$}
          child{
            node[draw,ellipse,align=center,scale=0.8]{uninstall\\shortcuts\\$(x_2)$}
            child{node[draw,rectangle,scale=0.8] {$-0.58$}
            edge from parent node[,scale=0.8,left] {no}
          }
          child{
            node[draw,rectangle,scale=0.8] {$0.06$}
            edge from parent node[,scale=0.8,right] {yes}
          }
          edge from parent node[,scale=0.8,left] {no}
        }
        child{
          node[draw,ellipse,align=center,scale=0.8]{install\\packages\\$(x_3)$}
	       child{node[draw,rectangle,scale=0.8] {$-0.19$}
          edge from parent node[,scale=0.8,left] {no}
        }
	    child{
          node[draw,rectangle,scale=0.8] {$0.34$}
          edge from parent node[,scale=0.8,right] {yes}
        }
        edge from parent node[,scale=0.8,right] {yes}
      }
      ;
      \end{scope}
    \end{tikzpicture}
  \end{minipage}
  \begin{minipage}{0,32\textwidth}
    \begin{tikzpicture}[->]
      \tikzstyle{level 1}=[sibling distance=22mm, level distance=20mm]
      \tikzstyle{level 2}=[sibling distance=10mm]
        \begin{scope}
        \node[scale=1.3] at (0,4) {$T_2$};
        \draw[rounded corners] (-2.6, -1.4) 
             rectangle (2.2 ,4.4)
        ;
      \end{scope}
      \begin{scope}[yshift=3cm]      
      \draw 
        node[draw,ellipse,align=center,scale=0.8]{read SMS\\$(x_4)$}
          child{
            node[draw,ellipse,align=center,scale=0.8]{write history\\\& bookmarks\\$(x_5)$}
            child{node[draw,rectangle,scale=0.8] {$-0.44$}
            edge from parent node[,scale=0.8,left] {no}
          }
          child{
            node[draw,rectangle,scale=0.8] {$0.23$}
            edge from parent node[,scale=0.8,right] {yes}
          }
          edge from parent node[,scale=0.8,left] {no}
        }
        child{
          node[draw,ellipse,align=center,scale=0.8]{read\\contacts\\$(x_6)$}
	       child{node[draw,rectangle,scale=0.8] {$0.29$}
          edge from parent node[,scale=0.8,left] {no}
        }
	    child{
          node[draw,rectangle,scale=0.8] {$-0.14$}
          edge from parent node[,scale=0.8,right] {yes}
        }
        edge from parent node[,scale=0.8,right] {yes}
      }
      ;
    \end{scope}
    \end{tikzpicture}
  \end{minipage}%
  \begin{minipage}{0,32\textwidth}
    \begin{tikzpicture}[->]
      \tikzstyle{level 1}=[sibling distance=20mm, level distance=20mm] 
      \tikzstyle{level 2}=[sibling distance=10mm]
        \begin{scope}
        \node[scale=1.3] at (0,4) {$T_3$};
        \draw[rounded corners] (-2.6, -1.4) 
             rectangle (2.2 ,4.4)
        ;
      \end{scope}
      \begin{scope}[yshift=3cm]         
      \draw 
        node[draw,ellipse,align=center,scale=0.8]{send SMS\\$(x_1)$}
          child{
            node[draw,ellipse,align=center,scale=0.8]{uninstall\\shortcuts\\$(x_2)$}
            child{node[draw,rectangle,scale=0.8] {$-0.39$}
            edge from parent node[,scale=0.8,left] {no}
          }
          child{
            node[draw,rectangle,scale=0.8] {$0.09$}
            edge from parent node[,scale=0.8,right] {yes}
          }
          edge from parent node[,scale=0.8,left] {no}
        }
        child{
          node[draw,ellipse,align=center,scale=0.8]{read\\contacts\\$(x_6)$}
	       child{node[draw,rectangle,scale=0.8] {$0.11$}
          edge from parent node[,scale=0.8,left] {no}
        }
	    child{
          node[draw,rectangle,scale=0.8] {$-0.17$}
          edge from parent node[,scale=0.8,right] {yes}
        }
        edge from parent node[,scale=0.8,right] {yes}
      }
      ;
      \end{scope}
    \end{tikzpicture}
  \end{minipage}
    \caption{A simple Android malware classifier trained on the DREBIN dataset.}
    \label{fig:drebin-model}
\end{figure}

\begin{eexample}[Prediction]
Consider an application that requests all of the permissions. 
By using the prediction function $f_{bin}$ from Definition~\ref{def:binary-classifier} 
together with the trees in Figure~\ref{fig:drebin-model}, we obtain the 
prediction 
$p_{\sigma}\big(0.34 + (-0.14) + (-0.17)\big) > 0.5 \Leftrightarrow 0.51 > 0.5$,
hence that particular application is classified as malware.
\end{eexample}
\subsection{Minimizing the Verbosity of Explanations}
\label{sec:minimize}
The ideas of minimizing the verbosity of explanations we explore in this paper 
originate from the problem of simplifying Boolean functions into prime 
implicants\cite{Quine52}, with later extensions to first-order logic for 
the purpose of abductive reasoning\cite{Marquis91}. More recently, 
these ideas have been applied to the explanation of predictions made by 
different types of machine learning 
models\cite{Shih18,Ignatiev19,La21,Ignatiev22,Audemard22}. These
related works are discussed further in Section~\ref{sec:related-works}. 
We now move on by defining the notion of a valid explanation.
\begin{definition}[Valid Explanation]
\label{def:valid-exp}
Let $f : \D^n \rightarrow \D$ be a function, $(x_1, \ldots, x_n)$ variables 
ranging over elements in $\D$, and $(c_1, \ldots, c_n) \in \D^n$ constant scalars. 
An explanation $E \subseteq \{(x_1, c_1), \ldots, (x_n, c_n)\}$ is valid with 
respect to a prediction $f(c_1, \ldots, c_n) \mapsto d$ iff
\begin{equation*}
  \Big(\bigwedge\limits_{(x_i, c_i) \in E} x_i = c_i \Big)
  \to f(x_1, \ldots, x_n) = d.
\end{equation*}
\end{definition}
\begin{eexample}[Valid Explanation]
Consider the classifier from Section~\ref{sec:cls-example} and an Android
application that requests all of the permissions, which is classified as 
malware. Clearly, the explanation 
$E_1 = \{(x_1, 1), (x_2, 1), (x_3, 1), (x_4, 1), (x_5, 1), (x_6, 1)\}$ which
contains all features is valid. Furthermore, we can see that a different
application that has the same permissions \textit{except} the ability to read 
contacts ($x_6$) is also classified as malware, i.e., 
\begin{equation*}
  \begin{alignedat}{4}
    &f_{bin}(1, 1, 1, 1, 1, 1; F) && = p_{\sigma}\big(0.34 + (-0.14) &&+ (-0.17)\big) &&> 0.5 \Leftrightarrow 0.51 > 0.5, \\
    &f_{bin}(1, 1, 1, 1, 1, 0; F) && = p_{\sigma}\big(0.34 + (0.29) &&+ (0.11)\big)   &&> 0.5 \Leftrightarrow 0.68 > 0.5.
  \end{alignedat}
\end{equation*}
Consequently, $E_2 = E_1 \backslash \{(x_6, 1)\} = \{(x_1, 1), (x_2, 1), (x_3, 1), (x_4, 1), (x_5, 1)\}$ is 
also a valid explanation, but with less redundant information than $E_1$.
\end{eexample}
In order to distinguish between several valid explanations, we define the 
notion of a minimal explanation.
\begin{definition}[Minimal Explanation]
\label{def:minimal}
An explanation $E$ that is valid with respect to a prediction 
$f(c_1, \ldots, c_n) \mapsto d$ is minimal iff removing elements from $E$
invalidates the explanation, i.e.,
\begin{equation*}
  \forall A \subset E, 
  \Big(\bigwedge\limits_{(x_i, c_i) \in A} x_i = c_i \Big)
    \not\to f(x_1, \ldots, x_n) = d.
\end{equation*}
\end{definition}
When formalized in propositional logic, minimal explanations are closely 
related to prime implicants, and thus are sometimes called
PI-explanations\cite{Shih18}. Other works sometimes call them
subset-minimal explanations\cite{Ignatiev19}, or sufficient
reasons\cite{Darwiche20}. 

In general, there may be several minimal explanations for a particular 
prediction. Depending on the application domain and target audience, different 
explanations may be preferable\cite{DARPA21}. In the medical 
domain, for example, some explanations may require that the target audience has
medical training in order to comprehend concepts captured by variables 
associated with an explanation, while other explanations may be more suitable 
for patients, albeit more verbose. In this work, we capture this need by 
providing the means to minimize the verbosity using different cost functions 
depending on the target audience.
\begin{definition}[Minimum Explanation]
\label{def:minimum}
A minimal explanation $E$ is minimum with respect to a cost function
$g: \powerset{\{1, \ldots, n\}} \rightarrow \R$ iff indices in $E$ minimize 
$g$, i.e.,
\[
   \{i: (x_i, c_i) \in E\} \in \argmin\limits_{\powerset{\{1, \ldots, n\}}}(g).
\]
\end{definition}
Note that there may be several minimum explanations for a particular 
prediction. We call the special case $E=\emptyset$ the empty explanation,
which is only valid for predictions made by a constant function. Some works
call minimum explanations minimal-cost explanations\cite{La21},
and when formalized in the satisfiability modulo theory, minimum explanations 
are closely related to minimum satisfying assignments. In this work, we focus
on explanations that are minimum with respect to the weighted sum 
$g(I; W) = \sum_{i \in I}w_i$, where $W =(w_1, \ldots, w_n) \in \R_{\ge 0}^n$
is an $n$-tuple with positive weights. Due to lack of domain knowledge in many
datasets used by researchers in experimental studies, however, all weights in 
our experiments are fixed to one. This is equivalent to the cost function 
$g(I) = |I|$, which is sometimes called cardinality-minimal 
explanations\cite{Ignatiev19}. 

\subsection{Exploring the Power Set Lattice of a Constraint System}
\label{def:expl-power-set}
In many areas of computer science, we are confronted with unsatisfiable 
constraint systems that need to be decomposed for pin-pointing causes of their
unsatisfiability. These decompositions may be represented by elements
in a power set lattice $(\powerset{C}, \subseteq)$, where $C$ is the set of
constraints put on the system, with $C=\emptyset$ representing the unconstrained 
system. Typically, we aim at finding elements adjacent to the frontier between
satisfiable and unsatisfiable subsets of $C$, where the ones on the upper side
(relative to the frontier) are called minimal unsatisfiable subsets (MUSes), 
and the lower ones are called maximal satisfiable subsets (MSSes).
\begin{definition}[Minimal Unsatisfiable Subset]
  \label{def:mus}
  A minimal unsatisfiable subset (MUS) of a set of constraints $C$ is a subset
  $S \subseteq C$ such that $S$ is unsatisfiable, and all proper subsets of $S$
  are satisfiable.
\end{definition}
\begin{definition}[Maximal Satisfiable Subset]
  \label{def:mss}
  A maximal satisfiable subset (MSS) of a set of constraints $C$ is a subset
  $S \subseteq C$ such that $S$ is satisfiable, and all proper supersets of $S$
  are unsatisfiable.
\end{definition}
In the context of linear programming, a MUS is often called an irreducible
infeasible subsystem (IIS), and a MSS is often called a maximal feasible subset 
(MFS).

With access to an oracle that can determine the satisfiability of subsets of a 
given constraint system, we can use a standard technique from linear programming
called the deletion filter\cite{Chinneck91} to shrink unsatisfiable 
subsets into MUSes, and similarly for growing satisfiable subsets into MSSes, 
as formalized in Figure~\ref{algo:shrink-and-grow}.

\begin{figure}
    \begin{subfigure}{0,47\textwidth}
    \begin{algorithmic}[1]
      \Function{Shrink}{$S$}
        \ForEach{$i \in S$}
          \If{\Not \Call{Oracle}{$S \backslash \{i\}$}}
            \Let{$S$}{$S \backslash \{i\}$}
          \EndIf
        \EndForEach
        \State \Return{$S$}
      \EndFunction
    \end{algorithmic}
    \end{subfigure}
    \begin{subfigure}{.5\textwidth}
    \begin{algorithmic}[1]
      \Function{Grow}{$S$}
        \ForEach{$i \in C \backslash S$}
          \If{\Call{Oracle}{$S \cup \{i\}$}}
            \Let{$S$}{$S \cup \{i\}$}
          \EndIf
        \EndForEach
        \State \Return{$S$}
      \EndFunction
    \end{algorithmic}
    \end{subfigure}\\

\caption{Shrink an unsatisfiable $S \subseteq C$ to a minimal unsatisfiable
             subset (MUS), or grow a satisfiable $S \subseteq C$ to a maximal satisfiable subset
            (MSS).}
                \label{algo:shrink-and-grow}
\end{figure}

We choose to encode each element $i \in C$ as the absence of the $i$-th variable
in an explanation (and thus the top element in the power set lattice encodes an
empty explanation), and thereby can derive a minimal explanation from the
complement of a MSS, corresponding to a minimal correction set.
\begin{definition}[Minimal Correction Set]
  \label{def:mcs}
  A minimal correction set (MCS) for a set of constraints $C$ is a subset
  $S \subseteq C$ such that $C \backslash S$ is a MSS.
\end{definition}

To determine the satisfiability of elements from a lattice with such an 
encoding, we use a valid explanation oracle in this paper.

\begin{definition}[Valid Explanation Oracle]
\label{def:valid-exp-oracle}
Let $f : \D^n \rightarrow \D$ be a function, $\bar{x} = (x_1, \ldots, x_n)$ variables 
ranging over elements in $\D$, $(c_1, \ldots, c_n) \in \D^n$ constants used 
in a prediction $p: f(c_1, \ldots, c_n) \mapsto d$, and 
$S \subseteq \{1, \ldots, n\}$ a set with elements that encode the 
\textit{absence} of variables (with given indices) in an explanation. A valid
explanation oracle is then defined as a procedure that decides the 
satisfiability of:
\begin{equation*}
  \forall \bar{x} \in \D^n, \Big(\bigwedge\limits_{i \in \{1, \ldots, n\} \backslash S} x_i = c_i \Big)
  \to f(\bar{x}) = d.
\end{equation*}
\end{definition}

Some other works\cite{Marques21,Ignatiev21} use a slightly different 
constraint system encoding to minimize explanations. In particular, these 
works encode elements in $C$ as the \textit{presence} of variables in an 
explanation, rather than their \textit{absence} as we do in this work. 
With that alternative encoding, one would use a negated valid explanation
oracle that determines the validity of a contrastive explanation (i.e., 
an explanation to a query on the form "why not") to find a MUS, which, 
according to the duality between contrastive and abductive 
explanations\cite{Ignatiev20}, is equivalent to a minimal abductive
explanation (i.e., an explanation to a query on the form "why").
We find our choice of encoding more natural for the problem of reducing the
verbosity of explanations, but the alternatives are equivalent with respect
to correctness.

\subsubsection{Finding a Minimum Explanation}
To find a minimum explanation, we explore the power set lattice in order to
determine the satisfiability of its elements in the following matter. Whenever 
a MSS is encountered, its corresponding MCS is computed, which captures all 
variable indices present in a minimal explanation, i.e., 
$E = \{(x_i, c_i): i \in S\}$, where $S$ is a MCS. 
When all MCSes have been considered, the minimum explanations with respect to
a cost function can be identified.


\begin{eexample}[Lattice Exploration]
Consider a simplification of the Android malware classifier from 
Section~\ref{sec:cls-example} restricted to a single tree $F = \{T_1\}$ as 
illustrated to the left in Figure~\ref{fig:lattice}, and the prediction
$p: f_{bin}(1, 1, 1; F) \mapsto 1$. To find a minimum explanation for $p$, 
we first define the set of constraints $C = \{1, 2, 3\}$, where each element 
$i \in C$ encodes the absence of the $i$-th variable in an explanation. We then
construct the power set lattice $(\powerset{C}, \subseteq)$, and assess the
satisfiability of its members by querying a valid explanation oracle. 
The right-hand side of Figure~\ref{fig:lattice} illustrates a Hasse diagram of
the constructed lattice, where each crossed-over member encodes the set of
\textit{absent} variables which would yield non-valid explanations, and each 
bold one is a MSS. As can be seen in that figure, the lattice contains two 
MSSes: $\{1\}$ and $\{2\}$, which correspond to the two MCSes: $\{1,3\}$ and 
$\{2, 3\}$, respectively. Consequently, the prediction $p$ has two minimal 
explanations: $\{(x_1, 1), (x_3, 1)\}$ and $\{(x_2, 1), (x_3, 1)\}$. 
With the cost function $g(I) = |I|$, both minimal explanations are also minimum.
\begin{figure}[ht]
  \begin{minipage}{0,49\textwidth}
    \center
    \begin{tikzpicture}[->]
      \tikzstyle{level 1}=[sibling distance=35mm, level distance=25mm]
      \tikzstyle{level 2}=[sibling distance=15mm]
        \begin{scope}
        \node[scale=1.3] at (0,4) {$T_1$};
      \end{scope}
      \begin{scope}[yshift=2.7cm]
      \draw 
        node[draw,ellipse,align=center,scale=1]{send SMS\\$(x_1)$}
          child{
            node[draw,ellipse,align=center,scale=1]{uninstall\\shortcuts\\$(x_2)$}
            child{node[draw,rectangle,scale=1] {$-0.58$}
            edge from parent node[,scale=1,left] {no}
          }
          child{
            node[draw,rectangle,scale=1] {$0.06$}
            edge from parent node[,scale=1,right] {yes}
          }
          edge from parent node[,scale=1,left] {no}
        }
        child{
          node[draw,ellipse,align=center,scale=1]{install\\packages\\$(x_3)$}
	       child{node[draw,rectangle,scale=1] {$-0.19$}
          edge from parent node[,scale=1,left] {no}
        }
	    child{
          node[draw,rectangle,scale=1] {$0.34$}
          edge from parent node[,scale=1,right] {yes}
        }
        edge from parent node[,scale=1,right] {yes}
      }
      ;
      \end{scope}
    \end{tikzpicture}

  \end{minipage}
  \begin{minipage}{0,49\textwidth}
  \center
  \[
  \def\arl{\ar@{-}}
    \xymatrix{& 
       \xcancel{\{1,2,3\}}\arl[dl]\arl[d]\arl[dr] & 
    \\ \xcancel{\{1,2\}}\arl[d]\arl[dr] & 
       \xcancel{\{1,3\}}\arl[dl]|\hole\arl[dr]|\hole &
       \xcancel{\{2,3\}}\arl[dl]\arl[d] 
    \\ \mathbf{\{1\}}\arl[dr] &
       \mathbf{\{2\}}\arl[d] & 
       \xcancel{\{3\}}\arl[dl]
    \\ & \{\} \\
  }
  \]
\end{minipage}

\caption{A simple Android malware classifier realized by a tree ensemble with a
         single tree (to the left), and a Hasse diagram (to the right) of the
         power set lattice of elements that can be removed from a valid 
         explanation to the prediction $p: f_{bin}(1, 1, 1; \{T_1\}) \mapsto 1$,
         where crossed-over elements yield non-valid explanations, and bold 
         elements yield minimal explanations.}
\label{fig:lattice}
\end{figure}
\end{eexample}

\subsubsection{The MARCO Algorithm}
MARCO (short for Mapping Regions of Constraints) is an algorithm that 
identifies the frontier between satisfiable and unsatisfiable subsets of 
constraints. It was originally presented in two independent
publications\cite{Liffiton13,Previti13}, and later unified by the same
set of authors\cite{Liffiton16}. 

The algorithm depends on a seed generator that yields unexplored elements from 
$\powerset{C}$, and maintains a state of explored elements with a set $M$. 
Normally, the number of elements in $M$ grows too large to reside in memory 
explicitly, hence one typically encodes them as an auxiliary formula with 
blocking constraints, which is solved using an off-the-shelf solver to extract
new seeds. In this work, we leverage a standard seed generator realized 
using propositional logic, where previously explored elements are blocked with
clauses in a Boolean CNF formula, and a SAT solver
is used to generate new ones, as formalized in Algorithm~\ref{algo:marco}.
\begin{algorithm}
    \caption{When equipped with a SAT solver and an oracle, the MARCO algorithm
             can enumerate all MUSes and MSSes of a set of constraints.}
    \label{algo:marco}
    \begin{algorithmic}[1]
      \Let{$M$}{$\emptyset$}
      \Comment{Initialize an empty Boolean CNF formula}
      \While{\Call{SAT}{$M$}}
        \Comment{Repeatedly explore elements in the power set lattice}
        \Let{$S$}{\Call{Solution}{$M$}}
        \Comment{Draw a seed from the lattice}
        \If{\Call{Oracle}{$S$}}
          \Comment{Check with oracle}
          \Let{$S$}{\Call{Grow}{$S$}}
          \Comment{Seed is satisfiable, grow to a MSS}
          \Let{$M$}{$M \cup \Big\{ \bigwedge\limits_{i \in C \backslash S} \phi_i$} \Big\}
          \Comment{Block subsets of the obtained MSS}
        \Else
          \Comment{Seed is unsatisfiable}
          \Let{$S$}{\Call{Shrink}{$S$}}
          \Comment{Shrink seed to a MUS}
          \Let{$M$}{$M \cup \Big\{ \bigwedge\limits_{i \in S} \neg\phi_i$} \Big\}
          \Comment{Block supersets of the obtained MUS}
        \EndIf
    \EndWhile
  \end{algorithmic}
\end{algorithm}

The algorithm starts by defining a CNF formula $M$, initialized with an
empty set of clauses (line~1). It then repeatedly checks the satisfiability 
of $M$ using a SAT solver (line~2). If $M$ is satisfiable, a solution $S$ 
is extracted from the solver, which becomes the new seed. The oracle is then 
queried for the satisfiability of $S$ (line~3). If the new seed is satisfiable, 
one climbs up in the lattice until a MSS is found (line~5). All subsets of the 
MSS are then blocked by appending a blocking clause to $M$ (line~6, where 
$\phi_i$ is a variable in the CNF formula). Otherwise, a MUS is found 
by climbing down the lattice, and all of its supersets are blocked (lines~8--9). 
This procedure is repeated until the SAT solver concludes that $M$ is 
unsatisfiable, in which case all MUSes and MSSes have been enumerated.

\subsection{Abstract Interpretation of Computer Programs}
\label{sec:absint}
Abstract interpretation is a framework introduced to facilitate sound and 
efficient reasoning about programs being analyzed by a compiler\cite{Cousot77}. 
The idea is to transform the source code of a program that computes values in a 
concrete domain into functions that operate in one or more abstract domains 
in which some analyses of interest are faster than in the corresponding concrete
domain, but potentially less precise. 

For example, to initiate the analysis of a program running with floating-point 
numbers to work within the domain of intervals, an abstraction function 
$\alpha: \PFP \rightarrow \I$ is used to map a set of values from the 
floating-point domain $\FP$ to values from the interval domain $\I$. 
Analogously, a concretization function $\gamma: \I \rightarrow \PFP$ is used to
map an interval to a set of floating-point numbers. Abstraction and 
concretization mappings that operate on elements from certain domains lead to a property 
called Galois connection that ensures sound reasoning with abstract 
interpretation\cite{Cousot77}.
\begin{definition}[Galois Connection]
\label{def:galois-connection}
Let $\alpha: \D \rightarrow \A$ and $\gamma: \A \rightarrow \PD$ be two 
monotone functions. A Galois connection $\PD \galois{\alpha}{\gamma} \A$ 
exist between the concrete domain $\D$ and the abstract domain $\A$ iff
\begin{gather*}\label{eq:galois-connection}
    \forall X \in \PD, X \subseteq \gamma(\alpha(X)), \\
    \forall \abs{x} \in \A, \abs{x} = \alpha(\gamma(\abs{x})).
  \end{gather*}
\end{definition}
To perform the analysis, the program is interpreted in the abstract domain by
evaluating sequences of abstract values, operators, and transformers.
\begin{definition}[Abstract Transformer]
\label{def:abstract-transformer}
An abstract transformer $\abs{f}: \A \rightarrow \A$ is an abstract counterpart
of a concrete function $f: \D \rightarrow \D$, but one that operates on elements
from the abstract domain. 
\end{definition}
When using abstract interpretation in a reasoning context, soundness can be 
ensured by proving that there exists a Galois connection between the concrete
and abstract domains, hence all computations performed in the abstract domain 
with the corresponding transformers are conservative.
\begin{definition}[Conservative Transformer]
\label{def:conservative-transformation}
Let $\alpha: \PD \rightarrow \A$ be an abstraction function, 
and $\gamma: \A \rightarrow \PD$ a concretization function.
An abstract transformer $\abs{f}: \A \rightarrow \A$ is conservative with
respect to a concrete function $f: \D \rightarrow \D$ iff
\begin{equation*}
\label{eq:conservative-transformation}
\forall X \in \PD, \forall x \in X, f(x) \in \gamma(\abs{f}(\alpha(X))).
\end{equation*}
\end{definition}
Abstract transformers for standard operators in concrete domains or sets
over such domains, e.g., addition and inclusion, can analogously be defined 
in a conservative manner. In this paper, we leverage an abstract interpreter 
that performs computations over elements from the interval domain.
\begin{definition}[Interval Domain]
\label{def:interval-domain}
The interval domain $\I$ contains abstract values $\abs{x}$ that capture sets 
of values from a concrete domain $\D$ using a range defined by a lower and upper
inclusive bound, i.e., $\abs{x} = [l,u]$, where $l, u \in \D, l \leq u$ are 
the lower and upper bound, respectively. Table~\ref{tbl:intdom} defines natural
abstraction and concretization functions, operators, and constants associated 
with the interval domain.
\begin{table*}[!t]%
  \centering %
  \caption{Abstract interpretation in the interval domain $\I$ over 
           values from the concrete domain $\D$, where $V \subseteq \D$, 
           $v, v_1, v_2, u_1, u_2 \in \D$, and
           $\abs{v} = [v_1, v_2], \abs{u} = [u_1, u_2] \in \I$.}%
\begin{tabular}{ll}
    \hline
    \textbf{Name}       & \textbf{Definition} \\
    \hline
    Abstraction  & $\alpha(V) = [\min V, \max V]$ \\
    Concretization & $\gamma([v_1, v_2]) = \{v : v_1 \leq v \leq v_2\}$ \\
    Top        & $\top = \alpha(\D)$ \\
    Bottom     & $\bot = \alpha(\emptyset)$\\
    Addition   & $\abs{v} + \abs{u} = [v_1 + u_1, v_2 + u_2]$ \\
    Join       & $\abs{v} \sqcup \abs{u} = [\min(v_1, u_1), \max(v_2, u_2)]$ \\
    Meet       & $\abs{v} \sqcap \abs{u} = 
                     \begin{cases} 
                        \bot \text{ if } \max(v_1,u_1) > \min(v_2,u_2),\\
                        [\max(v_1, u_1), \min(v_2, u_2)] \text{ otherwise} \\
                     \end{cases}
                 $ \\
    Greater than & $\abs{v} > \abs{u} = 
                     \begin{cases} 
                       [0, 0] \text{ if } \min(u_1, u_2) > \max(v_1, v_2), \\
                       [1, 1] \text{ if } \min(v_1, v_2) > \max(u_1, u_2), \\
                       [0, 1] \text{ otherwise}\\
                     \end{cases}
                 $ \\
    \hline
  \end{tabular}
  \label{tbl:intdom}
\end{table*}
\end{definition}
Abstract interpretation of programs with multiple variables in the interval
domain form hyperrectangles, in which case the domain is often referred to as
the Box domain.

\section{Related Works}
\label{sec:related-works}
The use of formal methods to prove the correctness of explanations for
predictions made by machine learning systems has been promoted fairly recently.
Shih~et~al.\cite{Shih18} compute minimal explanations for classifications made by
Bayesian networks. By compiling the prediction function of Bayesian networks
into an ordered (non-binary) decision diagram, they are able to compute
explanations efficiently. The computational challenges instead lie in the
compilation of Bayesian networks into decision diagrams, rather than the
computation of explanations\cite{Shih19}.

Ignatiev~et~al.\cite{Ignatiev19} demonstrate that off-the-shelf constraint solvers,
e.g., SMT and MILP solvers, can be used to compute both minimal and minimum
explanations if the prediction function can be represented within the used
reasoning engine. They assess scalability in terms of time taken to compute
explanations for predictions made by several neural networks trained on
different datasets. Through experiments, they demonstrate that the use of a
state-of-the-art MILP solver typically outperforms most SMT solvers, and
that computing explanations for models trained on high-dimensional inputs
is significantly more demanding than low-dimensional ones. They also
conclude that minimum explanations are typically less verbose than minimal
ones, but come at a significant computational cost. In this work, 
we develop a framework that embeds a reasoning engine tailored
specifically for tree ensembles to achieve improved runtime performance.
We also confirm that the trade-off between verbosity and computational
cost observed by Ignatiev~et~al. when explaining predictions made by
neural networks are also relevant for tree ensembles. 

La~Malfa~et~al.\cite{La21} compute explanations for neural networks, but use
reasoning engines designed specifically for neural networks. They compare two
different algorithms for computing a minimum explanation, both based on prior
work, but with several novel algorithmic improvements. One algorithm is based
on a branch-and-bound approach\cite{Dillig12}, and the other is based
on minimum hitting sets\cite{Ignatiev16}. In their 
Appendix\footnote{Available at https://arxiv.org/pdf/2105.03640.pdf}, 
they demonstrate that their improved branch-and-bound approach is typically
faster than the hitting sets approach. In this work, we use similar approaches
to compute minimum explanations, but for predictions made by tree ensembles, 
a model which has demonstrated greater predictive capabilities than neural
networks in several practical applications\cite{Shwartz22}.

Izza~and~Silva\cite{Izza21} use a SAT solver to compute minimal explanations for
non-trivial random forests (100 trees with depths ranging from three to eight).
They also show that this computational problem is $\text{D}^\text{P}$-complete,
i.e., the class of computations involving both NP-complete and coNP-complete 
sub-problems.
Around the same time, Boumazouza~et~al.\cite{Boumazouza21} also used a SAT solver to
compute minimal explanations for random forests, but with a different encoding.
They argue for the use of random forests as surrogates for models for which
no direct SAT-encoding exists.
Audemard~et~al.\cite{Audemard22} present the notion of majority reasons for the
purpose of explaining predictions made by random forests. These explanations
are significantly less computationally demanding to compute than minimal
explanations, but at the potential expense of some verbosity. They also show%
\footnote{Proof available at https://arxiv.org/pdf/2108.05276.pdf}
that the problem of computing a minimum explanation is $\Sigma_2^p$-complete,
i.e., NP-complete with a coNP oracle. In this work, we study both the problem
of computing a minimal explanation and a minimum explanation. 
Like the work by Audemard~et~al.\cite{Audemard22}, our focus is to reduce the
runtime needed to compute explanations, but here without compromising on verbosity.

Ignatiev~et~al.\cite{Ignatiev22} propose an approach that uses a MaxSAT-based oracle for 
computing minimal explanations of predictions made by tree ensembles. The 
approach is based on the application of the deletion 
filter\cite{Chinneck91}, which is implemented in a tool called XReason
and evaluated on gradient boosting machines trained on 21 different datasets.
They formulate the hard part of the MaxSAT encoding as a CNF formula, where
intervals are represented by auxiliary Boolean variables, while the validity
of an explanation is encoded as weighted soft clauses that can be optimized
to find a minimal explanation.
In this work, we also use a deletion filter, but with an oracle based on 
abstract interpretation which is designed specifically for tree ensembles. 
Compared to the MaxSAT approach, our oracle operates on an implicit DNF encoding
of relaxed constraints that captures input-output relations derived from a tree 
ensemble. The relaxed constraints are captured by abstract values from the 
interval domain, which are enumerated and iteratively tightened in an 
abstraction-refinement loop\cite{Tornblom19}. Furthermore, we tackle 
the computation of minimum explanations, and complete enumeration of all 
minimal explanations.

The MARCO algorithm\cite{Liffiton16} has been applied to a broad range
of problems, including for enumerating minimal explanations of predictions made
by decision lists\cite{Ignatiev21} and monotone
classifiers\cite{Marques21}. Compared to those works, we demonstrate 
that the pairing of the MARCO algorithm with an oracle formalized in the 
abstract interpretation framework can make the enumeration of all minimal 
explanations for several predictions made by non-trivial tree ensembles 
tractable. We also adapt MARCO for computing explanations that are minimum
with respect to the weighted sum $g(I; W) = \sum_{i \in I}w_i$, where 
$I \subseteq \{1, \ldots, n\}$ and $W = (w_1, \ldots, w_n) \in \R_{\ge 0}^n$,
demonstrating a speedup factor of 27 compared to the enumeration of all
minimal explanations.

A very recent work\cite{Izza22} tackles the problem of removing redundancy
in explanations for predictions made by decision trees using a MaxSAT approach. 
Our work applies to ensembles of decision trees, and aims for improved runtime 
performance by using abstract interpretation. Furthermore, we also aim for the
computation of explanations that are minimum with respect to a domain-specific 
cost function, as motivated by the comprehensive field guide on explainability
by Ras~et~al.\cite{Ras22}.

\section{An Oracle based on Abstract Interpretation}
\label{sec:oracle}
In this section, we formalize the inner workings of a valid explanation oracle
(Definition~\ref{def:valid-exp-oracle}), designed specifically for tree 
ensembles using the concept of abstract interpretation. First, we define 
abstract transformers (Section~\ref{sec:abs-transformers}), which are necessary
building blocks when constructing the oracle. Next, we define the abstract 
valid explanation oracle, together with an abstraction-refinement approach 
which ensures that abstractions that are too conservative are refined into 
more precise ones (Section~\ref{sec:abs-valid-oracle-def}). Finally, we 
assemble those building blocks into a sound and complete valid explanation 
oracle (Section~\ref{sec:abs-valid-oracle-construction}). For the purpose
of readability, proofs of lemmas and theorems stated in this 
section are given in Appendix~\ref{sec:proofs}.

Since the input dimension of a tree ensemble is typically different from 
its output dimension, we sometimes subscript abstraction and concretization 
functions to indicate which dimensionality they operate on, e.g., 
$\powerset{\D^n} \galois{\alpha_n}{\gamma_n} \A^n$, and we denote 
tuples of abstract values using capital letters, e.g., 
$\absvec{x} = (\abs{x}_1, \ldots, \abs{x}_n)$. Subscripted abstraction and 
concretization functions apply the corresponding non-subscripted function 
element-wise as follows:
\begin{equation*}
  \label{eq:subscript-notation}
  \begin{aligned}
    \absvec{x} &= \alpha_n(\{\bar{x}\}) \\
    &= \alpha_n(\{(x_1, \ldots, x_n)\}) \\ 
    &= (\alpha(\{x_1\}), \ldots, \alpha(\{x_n\})) \\
    &= (\abs{x}_1, \ldots, \abs{x}_n),
  \end{aligned}
\end{equation*}
where $\bar{x} \in \D^n$.
\subsection{Abstract Transformers}
\label{sec:abs-transformers}
We begin by defining the abstract transformers we later use to interpret 
tree-based classifiers: the decision tree transformer, the ensemble transformer, 
and the sigmoid transformer.

\begin{definition}[Tree Transformer]
\label{def:tree-transformer}
Let $T = \{(X_1, y_1), \ldots, (X_k, y_k)\}$ be a decision tree as 
defined by Definition~\ref{def:decision-tree}, and $\alpha$ an abstraction 
function. The decision tree transformer $\abs{t}$ is then defined as
\begin{equation*}
  \label{eq:tree-transformer}
  \abs{t}(\absvec{x}; T) = \alpha(\{y_i : (X_i,y_i) \in T, 
                           \absvec{x} \sqcap \alpha_n(X_i) \ne \bot\}).
\end{equation*}
\end{definition}
Given an abstract input tuple $\absvec{x}$, the transformer enumerates all pairs
$(X_i, y_i)$ from the decision tree $T$, and checks whether there are 
overlapping values between the abstraction of the input region $X_i$ and the 
values captured by $\absvec{x}$, in which case $y_i$ is included in 
the applied output abstraction. 

\begin{lemma}[Conservative Tree Transformer]
\label{lemma:tree}
The transformer $\abs{t}$ is conservative with respect to the prediction 
function $t$ (as defined by Definition~\ref{def:decision-tree}) if the 
abstraction function $\alpha$ forms a Galois connection with the used 
concretization function $\gamma$, i.e., that 
\[
  \forall \bar{x} \in \D^n, t(\bar{x}; T) \in \gamma(\abs{t}(\alpha_n(\{\bar{x}\})); T).
\]
\end{lemma}

Next, we define the ensemble transformer as the sum over a set of tree
transformations.
\begin{definition}[Ensemble Transformer]
\label{def:ensemble-transformer}
Let $F = (T_1, \ldots, T_B)$ be a tree ensemble as defined by
Definition~\ref{def:tree-ensemble}. The tree ensemble transformer is then 
defined as
\begin{equation*}
    \abs{f}(\absvec{x}; F) = \sum\limits_{i=1}^{B} \abs{t}(\absvec{x}; T_i).
\end{equation*}
\end{definition}
\begin{lemma}[Conservative Ensemble Transformer]
\label{lemma:ensemble}
The transformer $\abs{f}$ is conservative with respect to the prediction 
function $f$ (as defined by Definition~\ref{def:tree-ensemble}) 
if $\powerset{\D}\galois{\alpha}{\gamma}\A$ form a Galois connection, 
and the addition operator in the used abstract domain is conservative.
\end{lemma}
Since models trained by XGBoost on binary classification problems use the 
sigmoid function to make predictions, we need a transformer for that operation
as well.
\begin{definition}[Sigmoid Transformer]
Let $p_{\sigma}$ be the sigmoid function as defined by Definition~\ref{def:sigmoid}. 
The abstract sigmoid transformer $\abs{p}_{\sigma}$ is then defined as
\label{def:sigmoid-transformer}
\begin{equation*}
\abs{p}_{\sigma}(\abs{z}) = \alpha
\Big(\big\{
  p_{\sigma}(\min(\abs{z})), 
  \ldots,
  p_{\sigma}(\max(\abs{z}))
\big\}\Big)
\end{equation*}
\end{definition}
\begin{lemma}[Conservative Sigmoid Transformer]
\label{lemma:sigmoid}
The transformer $\abs{p}_{\sigma}$ is conservative with respect to the sigmoid
function $p_{\sigma}$.
\end{lemma}

With the basic building blocks of tree-based classifiers formalized as 
transformers, we can now define an abstract counter-part to the
binary classifier $f_{bin}$ introduced in Section~\ref{sec:classification}.
\begin{definition}[Abstract Binary Classifier]
\label{def:abs-bin-classifier}
Let $F$ be a tree ensemble as defined by Definition~\ref{def:binary-classifier}. 
The abstract binary classifier $\abs{f}_{bin}$ is then defined as
\begin{equation*}
    \abs{f}_{bin}(\absvec{x}; F) = \abs{p}_{\sigma}(\abs{f}(\absvec{x}; F)) > \alpha(\{0.5\}).
\end{equation*}
\end{definition}
\begin{lemma}
\label{lemma:bin-classifier}
The abstract binary classifier $\abs{f}_{bin}$ is conservative with respect to
the binary classifier $f_{bin}$ (as defined by Definition~\ref{def:binary-classifier})
if the transformer for the operator $>$ is conservative.
\end{lemma}

\begin{eexample}[Abstract Prediction]
Consider an Android malware classifier with a single tree from 
Section~\ref{sec:cls-example}, here illustrated by 
Figure~\ref{fig:tree-transformer-example} together with its formalized 
input-output relation $T_1$ (according to Definition~\ref{def:decision-tree}). 
\begin{figure}[ht]
  \center
  \begin{minipage}{0,44\textwidth}    
    \begin{tikzpicture}[->]
      \tikzstyle{level 1}=[sibling distance=35mm, level distance=23mm]
      \tikzstyle{level 2}=[sibling distance=15mm]
      \begin{scope}[yshift=3cm]
      \draw 
        node[draw,ellipse,align=center,scale=0.88]{send SMS\\$(x_1)$}
          child{
            node[draw,ellipse,align=center,scale=0.88]{uninstall\\shortcuts\\$(x_2)$}
            child{node[draw,rectangle,scale=0.88] {$-0.58$}
            edge from parent node[,scale=0.88,left] {no}
          }
          child{
            node[draw,rectangle,scale=0.88] {$0.06$}
            edge from parent node[,scale=0.88,right] {yes}
          }
          edge from parent node[,scale=0.88,left] {no}
        }
        child{
          node[draw,ellipse,align=center,scale=0.88]{install\\packages\\$(x_3)$}
	       child{node[draw,rectangle,scale=0.88] {$-0.19$}
          edge from parent node[,scale=0.88,left] {no}
        }
	    child{
          node[draw,rectangle,scale=0.88] {$0.34$}
          edge from parent node[,scale=0.88,right] {yes}
        }
        edge from parent node[,scale=0.88,right] {yes}
      }
      ;
      \end{scope}
    \end{tikzpicture}
  \end{minipage}
  \begin{minipage}{0,44\textwidth}

    \begin{equation*}
      T_1 = 
        \left\{\begin{alignedat}{3}
          &\big(\{(0, 0, 0), (0, 0, 1)\},& -0.58 & \big), \\
          &\big(\{(0, 1, 0), (0, 1, 1)\},& 0.06  & \big), \\
          &\big(\{(1, 0, 0), (1, 1, 0)\},& -0.19 & \big), \\
          &\big(\{(1, 0, 1), (1, 1, 1)\},& 0.34  & \big)
        \end{alignedat}\right\}
    \end{equation*}
    \end{minipage}
    \caption{A simple decision tree trained by XGBoost on the DREBIN 
             dataset (to the left), together with its formalized input-output
             relation (to the right).}
    \label{fig:tree-transformer-example}
\end{figure}

Now, suppose we are interested in the input-output relation of the classifier 
for all applications with the permission to uninstall shortcuts and 
install packages ($x_2$ and $x_3$). We can capture these inputs with a single 
element from the interval domain $\I^3$ as
\[
\abs{X} = \alpha_3\big(\{(0, 1, 1), (1, 1, 1)\}\big) = \{([0, 1], [1, 1], [1, 1])\}.
\]
Next, we use the transformer $\abs{t}$ to compute an 
abstract output $\abs{t}(\absvec{x}; T_1) = [0.06, 0.34]$, and conclude that 
all applications captured by $\absvec{x}$ are classified as malware since
$0.5 < p_{\sigma}(0.06) < p_{\sigma}(0.34) \Leftrightarrow 0.5 < 0.51 < 0.58$.
\end{eexample}

\subsection{Abstract Valid Explanation Oracle}
\label{sec:abs-valid-oracle-def}
Recall the problem we are tackling in this section, i.e., given a prediction
$p$ and an explanation $E$, determine if $E$ is valid with respect to $p$. 
\begin{definition}[Abstract Valid Explanation Oracle for Binary Classifications]
\label{def:abs-expl-oracle}
Let $p: f_{bin}(c_1, \ldots, c_n; F) \mapsto d$ be a prediction, 
$S \subseteq \{1, \ldots, n\}$ a set with elements that encode the 
absence of variables with given indices in an explanation (as defined
by Definition~\ref{def:valid-exp-oracle}), and 
$\absvec{X} = (\abs{x}_1, \ldots, \abs{x}_n)$ an abstract input tuple that
captures all possible combinations of assignments to those absent variables, 
where the $i$-th element in $\absvec{X}$ is defined as
\begin{equation*}
  \abs{x}_i = 
  \begin{cases} 
    \top & \text{if } i \in S, \\
    \alpha({\{c_i\}}) & \text{otherwise}.
  \end{cases}
\end{equation*}
An abstract valid explanation oracle $v_{bin}$ for binary classifications may 
then be defined as
\begin{equation*}
  v_{bin}(\absvec{X}, d; F) = 
  \begin{cases} 
    Pass & \text{if } \{d\} = \gamma(\abs{f}_{bin}(\absvec{X}; F)), \\
    Fail & \text{if } d \not\in \gamma(\abs{f}_{bin}(\absvec{X}; F)), \\
    Unsure & \text{otherwise}.
  \end{cases}
\end{equation*}
\end{definition}
\begin{theorem}[Soundness]
\label{theo:sound}
Given an explanation (as used within Definition~\ref{def:abs-expl-oracle}),
the abstract explanation oracle $v_{bin}$ that uses the transformer 
$\abs{f}_{bin}$ is sound, i.e., whenever $v_{bin}$ returns 
$Pass$, the explanation is valid, and whenever $v_{bin}$ returns $Fail$, 
it is not valid.
\end{theorem}

In general, abstract interpretation is not complete, i.e., abstract transformers
may yield abstractions that are too conservative in order to provide conclusive
query responses when used in a reasoning engine, i.e., when $v_{bin}$ returns
$Unsure$. In our case where tree ensembles only contain univariate and linear
decision rules, however, there exists algorithms designed specifically for
abstract interpretation of tree ensembles that refine abstractions that are 
too conservative into several more concise ones, eventually leading to precise 
abstractions\cite{Tornblom19,Ranzato20}. 
In this work, we leverage one of these algorithms, implemented in the tool 
suite VoTE\cite{Tornblom19}, to construct an oracle that is both sound
and complete. VoTE has demonstrated great performance in terms of runtime and
memory usage\cite{Tornblom21}, and provides a modular property checking 
interface realized as a recursive higher-order function. More specifically, 
VoTE accepts three parameters as input: a tree ensemble $F$, a tuple of 
intervals $\absvec{x} \in \I^n$ with $\bot \not\in \absvec{x}$, and a 
property checker $pc: \I^n \times \I \rightarrow \{Pass,Fail,Unsure\}$, as 
formalized by Algorithm~\ref{algo:VoTE}. The algorithm also accepts an 
auxiliary parameter $R \subseteq F$ that keeps track of refinement steps
during recursion, and is initialized to the empty set.

\begin{algorithm}[ht]
  \caption{The abstraction-refinement algorithm implemented in VoTE, where
           $F$ is a tree ensemble, $\absvec{x}$ a tuple of intervals that capture
           points in the input domain, $pc$ a property checker, and $R$ a
           state-tracking auxiliary parameter initialized to $\emptyset$.}
  \label{algo:VoTE}
  \begin{algorithmic}[1]
    \Function{VoTE}{$F, \absvec{x}, pc, R=\emptyset$}
      \Let{$o$}{$pc(\absvec{x}, \abs{f}(\absvec{x}; F))$}
      \Comment{Compute outcome of property checker}
      \If{$o \ne Unsure$}
        \State \Return $o$
        \Comment{Property checker is conclusive with this abstraction}
      \ElsIf{$R = F$}
        \State \Return $o$
        \Comment{No more refinements necessary}
      \Else
        \State {$T \in F \backslash R$}
        \Comment{Select a tree not in $R$}
        \ForEach{$(X_i, y_i) \in T$}
          \Comment{Enumerate leaves in the selected tree}
          \Let{$\absvec{x}_i$}{$\absvec{x} \sqcap \alpha_n(X_i)$}
          \Comment{Refine abstraction}
          \Let{$o$}{\Call{VoTE}{$F, \absvec{x}_i, pc, R \cup \{T\}$}}
          \Comment{Recursively check the refined abstraction}
          \If{$o \ne Pass$}
            \State \Return $o$
            \Comment{Stop enumerating if outcome is $Fail$ or $Unsure$}
          \EndIf
        \EndForEach
        \State \Return $Pass$
        \Comment{Property is satisfied for all inputs captured by $\absvec{x}$}
      \EndIf
    \EndFunction
  \end{algorithmic}
\end{algorithm}

The algorithm starts by computing a conservative output approximation 
$\abs{f}(\absvec{x}; F)$, which is then checked by the given property checker 
$pc$ (line~2). If the property checker is conclusive, or if there are no more
refinements necessary (see the following Lemma~\ref{lemma:precise} and its 
proof in Appendix~\ref{sec:proofs} why this is the case when $R = F$), the 
outcome of $pc$ is returned (lines~3--6). Otherwise, the algorithm picks an 
arbitrary tree $T$ from the 
ensemble $F$ that has not been picked before (line~8), and refines $\absvec{x}$ 
into a partition with abstract input tuples $\absvec{x}_1, \ldots, \absvec{x}_k$
(lines~9--15). For each refined abstraction $\absvec{x}_i$, the algorithm is
invoked recursively (line~11). Whenever a recursive invocation yields an 
outcome that does not guarantee satisfaction with the property checked by $pc$, 
i.e., whenever $pc$ returns $Fail$ or $Unsure$, the computed outcome is 
returned. When the tree ensemble satisfies the checked property for all input
values captured by $\absvec{x}$, the algorithm returns $Pass$ (line~16).

\begin{lemma}[Precise Abstraction Refinement]
\label{lemma:precise}
Given a tuple of abstractions $\absvec{x}$ with $\bot \not\in \absvec{X}$,
and a property checker $pc$ that always returns $Unsure$, 
Algorithm~\ref{algo:VoTE} refines $\absvec{x}$ into a partition with more 
concise abstract input tuples $\absvec{x}_1, \ldots, \absvec{x}_k$, which 
eventually leads to precise output abstractions, i.e.,
$\forall i \in \{1, \ldots, k\}, |\gamma(\abs{f}(\absvec{x}_i; F))| = 1$.
\end{lemma}

\subsubsection{The Construction of a Valid Explanation Oracle}
\label{sec:abs-valid-oracle-construction}
We now combine VoTE with a property checker that realizes $v_{bin}$, as 
formalized in Algorithm~\ref{algo:oracle}. That algorithm takes as input a 
tree ensemble $F$, a tuple $\absvec{X}$ of abstract input values constructed 
from an explanation $E$ (according to Definition~\ref{def:abs-expl-oracle}), 
and the predicted label $d$. 
The algorithm returns $True$ if all input values captured by $\absvec{X}$ are 
mapped to the label $d$ by the classifier $f_{bin}$, in which case $E$ is 
valid with respect to $p$. Otherwise, there is some input value $\bar{x}$ 
captured by $\absvec{X}$ such that $f_{bin}(\bar{x}) \ne d$, in which case 
$E$ is not valid with respect to $p$.

\begin{algorithm}
  \caption{A valid explanation oracle that checks if the tree ensemble $F$ 
           maps all points captured by the abstract input tuple 
           $\absvec{x} = (\abs{x}_1, \ldots, \abs{x}_n)$ to the label 
           $d \in \{0, 1\}$.}
  \label{algo:oracle}
  \begin{algorithmic}[1]
    \Function{Is\_Valid}{$F, \absvec{x}, d$}
      \Function{$pc$}{$\absvec{x}_p, \abs{y}_p$}
      \Comment{Define a property checker with a closure that binds $d$}
        \Let{$D$}{$\gamma\big(\abs{p}_{\sigma}(\abs{y}_p) > \alpha(\{0.5\})\big)$}
        \Comment{Equivalent to $\gamma\big(\abs{f}_{bin}(\absvec{x}; F)\big)$ since $\abs{y} = \abs{f}(\absvec{x}; F)$}
        \If{$|D| \ne 1$}
          \State \Return $Unsure$
          \Comment{Predicted output may be either $0$ or $1$}
        \ElsIf{$d \in D$}
          \State \Return $Pass$
          \Comment{Explanation is valid}
        \EndIf
        \State \Return $Fail$
        \Comment{Explanation is not valid}
      \EndFunction
      \State \Return \Call{VoTE}{$F, \absvec{x}, pc$} = $Pass$
      \Comment{Return $True$ if VoTE returns $Pass$}
    \EndFunction
  \end{algorithmic}
\end{algorithm}

Algorithm~\ref{algo:oracle} essentially calls VoTE (line~11) to systematically 
refine the given $\absvec{X}$ into a partition with precise abstractions 
$\absvec{X}_1, \ldots, \absvec{X}_k$, which are enumerated and checked with
the oracle $v_{bin}$. To fit within the VoTE framework, $v_{bin}$ is realized 
as a property checker $pc$ (lines~2-10), which takes as input a pair 
$(\absvec{x}_p, \abs{y}_p)$, and is invoked by VoTE such that
$\abs{y}_p = \abs{f}(\absvec{x}_p; F)$ (see line~2 in Algorithm~\ref{algo:VoTE}). 
Consequently, $pc$ may complete the validity check by computing 
$\abs{p}_{\sigma}(\abs{y}) < \alpha(\{0.5\})$, followed by a concretization 
into a set of potential labels $D$ (line~3). If $D$ contains multiple labels, 
$pc$ returns $Unsure$ (line~5), which causes VoTE to refine $\absvec{X}_p$ 
(see lines~8--15 in Algorithm~\ref{algo:VoTE}). If $D$ contains exactly the 
label $d$, the property checker returns $Pass$ (line~7), which causes VoTE to
continue processing the remaining elements in the partition of $\absvec{X}$. 
Otherwise, the property checker returns $Fail$ (line~9), which causes the 
call to VoTE on line~11 to return $Fail$.
\begin{theorem}[Correctness]
\label{theo:correct}
Algorithm~\ref{algo:oracle} is a sound and complete valid explanation 
oracle.
\end{theorem}

\begin{eexample}[Validity of an Explanation]
Again consider the classifier from Section~\ref{sec:cls-example}, and an Android
application that requests all of the permissions (which is classified as 
malware). Now, suppose we would like to check the validity of the explanation
that contains all variables except those encoding the permissions to install 
packages and read contacts ($x_3$ and $x_6$), i.e., that all applications 
captured by $\absvec{X} = ([1, 1], [1, 1], \top, [1, 1], [1, 1], \top)$ 
are classified as malware. We invoke the oracle with the arguments 
$(F, \absvec{X}, 1)$, and observe the abstraction-refinement approach
in each recursive invocation. The rows in Table~\ref{tbl:absref-example}
lists snapshots of values whenever Algorithm~\ref{algo:VoTE}
is invoked, where $R$ is the auxiliary state variable, $\absvec{X}$ the 
input region subject to analysis, $\abs{t}(\absvec{X}; T_i)$ the output from 
the $i$-th tree, $\abs{f}(\absvec{X}; F)$ the output from the tree ensemble, 
and $o$ the outcome of the property checker $pc$ (defined on line~2 in 
Algorithm~\ref{algo:oracle}).

\begin{table*}[!t]%
  \centering
  \caption{Snapshots of values computed by Algorithm~\ref{algo:oracle} in
           each recursive invocation when checking the validity of an
           explanation, where snapshots taken at different recursive depths
           are separated by a dashed line.}
  \begin{tabular}{c|c|c|c|c|c|c}
    \hline
    $R$ & $\absvec{X}$ & $\abs{t}(\absvec{X}; T_1)$ 
                       & $\abs{t}(\absvec{X}; T_2)$ 
                       & $\abs{t}(\absvec{X}; T_3)$ 
                       & $\abs{f}(\absvec{X}; F)$ 
                       & $o$
    \\
    \hline
    
    \rule{0pt}{\normalbaselineskip}
    $\emptyset$ & $([1,1], [1,1], \top, [1,1], [1,1], \top)$ & $[-0.19, 0.34]$ & $[-0.14, 0.29]$ & $[-0.17, 0.11]$ & $[-0.49, 0.73]$ & \qmark \\
    \hdashline
    
    \rule{0pt}{\normalbaselineskip}
    $\{T_1\}$ & $([1,1], [1,1], [1,1], [1,1], [1,1], \top)$ & $[0.34,0.34]$ & $[-0.14, 0.29]$ & $[-0.17, 0.11]$ & $[0.03, 0.73]$ & \cmark \\

    \rule{0pt}{\normalbaselineskip}
    $\{T_1\}$ & $([1,1], [1,1], [0,0], [1,1], [1,1], \top)$ & $[-0.19,-0.19]$ & $[-0.14, 0.29]$ & $[-0.17, 0.11]$ & $[-0.49, 0.21]$ & \qmark \\
    \hdashline

    \rule{0pt}{\normalbaselineskip}
    $\{T_1, T_2\}$ & $([1,1], [1,1], [0,0], [1,1], [1,1], [1,1])$ & $[-0.19,-0.19]$ & $[-0.14,-0.14]$ & $[-0.17,-0.17]$ & $[-0.49,-0.49]$ & \xmark \\
    
    \hline
  \end{tabular}
  \label{tbl:absref-example}
\end{table*}

In the first invocation (where $R=\emptyset$), the abstract input region 
$\absvec{X}$ is too conservative, leading a an inconclusive outcome. The 
algorithm then uses the first tree ($T_1$) to refine $\absvec{X}$ into two 
adjacent abstract input regions, with each one being used as input to two 
sequential calls to VoTE (now with $R=\{T_1\}$). In the first of these cases, 
all inputs captured by the refined region are classified as malware, leading 
to a conclusive outcome. In the second case, however, the outcome is still 
inconclusive, and the algorithm uses the second tree ($T_2$) to further refine 
the remaining input region. In the last recursive invocation ($R=\{T_1, T_2\}$), 
$\absvec{X}$ captures a single value, which is classified as benign. 
Consequently, the explanation that contains all variables except those encoding
the permissions to install packages and read contacts ($x_3$ and $x_6$) is not 
valid for the given prediction.
\end{eexample}

\section{Algorithms for Minimizing Explanations}
\label{sec:algos}
In this section, we formalize two algorithms that use the oracle from 
Section~\ref{sec:oracle} to minimize explanations: one for computing minimal 
explanations, and one for computing explanations that are minimum with respect
to the weighted sum of the indices of variables present in the explanation.
Proofs of correctness for these algorithms are given in Appendix~\ref{sec:proofs}.
The problem of determining the validity of an explanation is in the set of
coNP problems\cite{Izza21}, and the problem of finding a minimal 
explanation is $\text{D}^\text{P}$-complete\cite{Izza21}. 
The problem of finding a minimum explanation, however, is in the set of 
$\Sigma_2^p$-complete problems\cite{Audemard22}. In Section~\ref{sec:perf} we 
show that with access to a highly efficient oracle, however, there are
applications in the context of current published benchmarks where the 
computation of minimum explanations is both desirable and tractable.

\subsection{Computing a Minimal Explanation}
We base our approach for computing minimal explanations on the deletion 
filter\cite{Chinneck91}, here formalized as Algorithm~\ref{algo:minimal} 
in the context of abstract interpretation. Our algorithm takes as input a 
tree ensemble $F$, the set of variables referenced by the tree ensemble 
$V_f$ (obtained through Definition \ref{def:var-ref-tree} and
\ref{def:var-ref-ensemble}), 
the input values used in a given prediction $c_1, \ldots, c_n$, the 
predicted label $d$, and returns all indices of the variables present in 
some minimal explanation.
\begin{algorithm}
    \caption{Compute a minimal explanation for a given tree ensemble $F$ 
      that references variables with indices in $V_f$, and an input
      tuple ($c_1, \ldots, c_n$) classified as label $d$.}
    \label{algo:minimal}
    \begin{algorithmic}[1]
    \Function{Minimal\_Explanation}{$F, V_f, c_1, \ldots, c_n, d$}
    \Let{$S$}{$\emptyset$}
    \Comment{Initialize set of variable indices to delete}
    \Let{$\abs{x}_1, \ldots, \abs{x}_n$}{$\alpha(\{c_1\}), \ldots, \alpha(\{c_n\})$}
    \Comment{Initialize abstractions of the input values}
    \ForEach{$i \in \{1 \ldots n\}$}
        \Let{$\abs{x}_i$}{$\top$}
        \Comment{Relax $i$-th abstraction}
        \If{$i \in V_f$}
          \Comment{No need to check unused input variables}
          \If{\Call{Is\_Valid}{$F, (\abs{x}_1, \ldots, \abs{x}_n), d$}}
          \Let{$S$}{$S \cup \{i\}$}
          \Comment{The $i$-th variable can be removed}
          \Else
          \Let{$\abs{x}_i$}{$\alpha(\{c_i\})$}
          \Comment{The $i$-th variable affects the outcome, restore it}
          \EndIf
        \EndIf
    \EndForEach
    \State \Return $V_f \backslash S$
    \Comment{Return indices in a minimal explanation}
    \EndFunction
  \end{algorithmic}
\end{algorithm}

First, we initialize a set of variable indices to delete as the empty set
(line~2), and a tuple of abstractions to capture precisely the concrete input
tuple used in the prediction subject to explanation (line~3). Next, we pick
successive abstractions $\abs{x}_i$, and widen the selected one to include all
possible assignments to that particular variable (line~5). On line~6, we check
if the index of the selected abstraction is in the set of variables referenced
by the tree ensemble ($V_f$). If that is the case, we use
Algorithm~\ref{algo:oracle} as an oracle to check if all values captured by 
the (widened) abstract input tuple $(\abs{x}_1, \ldots, \abs{x}_n)$ map to the 
label $d$ (line~7). If the oracle decides that this is the case, we add the
index of the selected (widened) abstraction to the set of indices to delete 
(line~8). Otherwise, the value of that specific input variable was relevant
for the prediction, and $\abs{x}_i$ is restored to only include the value used
during prediction (line~10). This process is then repeated over all input 
dimensions (lines~4--13). Finally, the set of variable indices that indeed
influence the prediction is returned (line~14).

\begin{proposition}[Correctness]
\label{prop:minimal}
Let $F$ be a tree ensemble, and $p: f_{bin}(c_1, \ldots, c_n; F) \mapsto d$
a prediction. The set returned by Algorithm~\ref{algo:minimal} contains the 
indices of a minimal explanation for $p$.
\end{proposition}

\subsection{Computing a Minimum Explanation}
As mentioned in Section~\ref{sec:minimize}, there may be several minimal 
explanations to choose from for a particular prediction, and by leveraging a
domain-specific cost-function, minimum explanations may be tailored specifically
for their target audience. Here, we present an algorithm we call m-MARCO that
can compute explanations that are minimum with respect to the cost function
$g(I; W) = \sum_{i \in I}w_i$, where $I \subseteq \{1, \ldots, n\}$ are 
indices of variables present in the explanation, and
$W = (w_1, \ldots, w_n) \in \R_{\ge 0}^n$ are positive weights. The algorithm 
accepts the same input parameters as Algorithm~\ref{algo:minimal}, i.e., a tree 
ensemble $F$, the set of variables referenced by the tree ensemble $V_f$, 
the input values used in the prediction $c_1, \ldots, c_n$, the predicted 
label $d$, and it returns the indices of the variables present in some 
minimum explanation. There is also an optional parameter 
$W = (w_1, \ldots, w_n)$ with weights for the cost function that defaults
to $w_i = 1$. Note that the default configuration of the cost function is 
equivalent to using $g(I) = |I|$, in which case m-MARCO computes an explanation
of minimum size.

Recall from Section~\ref{def:expl-power-set} that we can use MARCO to enumerate
MSSes from a power set lattice of a constraint system. By encoding elements in
the lattice as sets of variable indices that are absent from an explanation, we
can derive minimal explanations from MSSes by computing their corresponding 
MCSes. Each MCS then encodes the variable indices present in a minimal 
explanation. With m-MARCO (Algorithm~\ref{algo:minimum-marco}), we present an 
adaptation of the standard MARCO approach\cite{Liffiton16} that uses this 
encoding together with the oracle formalized by Algorithm~~\ref{algo:oracle}.
The adaptations are indicated by line numbers with a gray background in 
Algorithm~\ref{algo:minimum-marco}, where the key differences are defined on 
lines 13--14 and 18--20. In particular, m-MARCO is able to block some of the
unexplored seeds where the corresponding explanation is more verbose than any
of the previously enumerated ones.

The algorithm takes as input a tree ensemble $F$, the set of variables
referenced by the ensemble $V_f$, the inputs used in the prediction
$(c_1, \ldots, c_n)$, the predicted label $d$, and the weights for each
feature $W$ (which defaults to $\{1\}^n$), and depends on a SAT solver that
accepts formulas with cardinality constraints.

\begin{algorithm}[ht]
    \caption{The m-MARCO approach computes an explanation that is minimum
             with respect to a cost function $g$ for a tree ensemble $F$,
             where $V_f$ is the set of variables referenced by $F$
             (see Definition~\ref{def:var-ref-ensemble}), $(c_1, \ldots, c_n)$
             the input tuple, and $d$ the predicted class.}
    \label{algo:minimum-marco}
    \begin{algorithmic}[1]
      \boldnext
      \Function{Minimum\_Explanation}{$F, V_f, c_1, \ldots, c_n, d, W=\{1\}^n$}
        \boldnext
        \Function{Oracle}{$S$}
          \Comment{Define an oracle interface for MARCO}
          \boldnext
          \Let{$\abs{x}_1, \ldots, \abs{x}_n$}{$\top, \ldots, \top$}
          \Comment{Initialize all abstractions}
          \boldnext
          \ForEach{$i \in V_f \backslash S$}
          \Comment{Compute MCS}
            \boldnext
            \Let{$\abs{x}_i$}{$\alpha(\{c_i\})$}
            \Comment{Assign value to the $i$-th variable}
            \boldnext
          \EndForEach
          \boldnext
          \State \Return{\Call{Is\_Valid}{$F, (\abs{x}_1, \ldots, \abs{x}_n), d$}}
          \Comment{Query the oracle}
          \boldnext
        \EndFunction
      \boldnext
      \Let{$S_{cur}$}{$\emptyset$}
      \Comment{Initialize a set of variable indices to remove}
      \Let{$M$}{$\emptyset$}
      \Comment{Initialize an empty Boolean CNF formula}
      \While{\Call{SAT}{$M$}}
        \Comment{Repeatedly explore elements in the power set lattice}
        \Let{$S$}{\Call{Solution}{$M$}}
        \Comment{Draw a seed from the lattice}
        \boldnext
        \If{$g(V_f \backslash S_{cur}; W) \le g(V_f \backslash S; W)$}
          \boldnext
          \Let{$M$}{$M \cup \Big\{ \bigwedge\limits_{i \in V_f \backslash S} \phi_i$} \Big\}
          \Comment{Block subsets of seed}
        \ElsIf{\Call{Oracle}{$S$}}
          \Comment{Check with oracle}
          \Let{$S$}{\Call{Grow}{$S$}}
          \Let{$M$}{$M \cup \Big\{ \bigwedge\limits_{i \in V_f \backslash S} \phi_i$} \Big\}
          \Comment{Block subsets of MSS}
          \boldnext
          \Let{$S_{cur}$}{$S$}
          \boldnext
          \Let{$k$}{$\min\limits_{S \subseteq V_f} \Big\{|S|: g(V_f \backslash S_{cur}; W) > g(V_f \backslash S; W) \Big\}$}
          \Comment{\parbox[t]{.3\linewidth}{Find the cardinality of the smallest seed that removes more verbosity than $S_{cur}$}}
        
          \boldnext
          \Let{$M$}{$M \cup \Big\{k \ge \sum\limits_{i \in V_f} \phi_i$} \Big\}
          \Comment{Block seeds with cardinality lower than $k$}
        \Else
          \Comment{Seed is unsatisfiable}
          \Let{$S$}{\Call{Shrink}{$S$}}
          \Comment{Shrink seed to a MUS}
          \Let{$M$}{$M \cup \Big\{ \bigwedge\limits_{i \in S} \neg\phi_i$} \Big\}
          \Comment{Block supersets of MUS}
        \EndIf
    \EndWhile
    \boldnext
    \State \Return{$V_f \backslash S_{cur}$}
    \Comment{Return variable indices in a minimum explanation}
    \boldnext
    \EndFunction
  \end{algorithmic}
\end{algorithm}

First, we define a function that acts as a front-facing oracle interface for
the MARCO algorithm that accepts as input a set of variable indices $S$ to 
remove from an explanation (lines~2--8). That function starts by initializing
a tuple of abstract values $(\abs{x}_1, \ldots, \abs{x}_n)$ which capture all
possible inputs to the tree ensemble (line~3). It then enumerates all 
variables referenced by the tree ensemble, except those to exclude from the 
explanation, and tightens their abstractions to only include the value each 
enumerated variable had during the prediction (lines~4--6). Finally, the 
function invokes the underlying oracle (line~7).

With an oracle defined with an interface compatible with the MARCO algorithm, we
initialize a set $S_{cur}$ of variable indices to remove to the empty set of
indices (line~9), and a CNF formula $M$ to the empty set of clauses (line~10). 
We then repeatedly check the satisfiability of $M$ using a SAT solver (line~11). 
If $M$ is satisfiable, we extract a solution from the solver, which becomes our
new seed (line~12). We then check if removing variables with indices in the 
seed yields a more verbose explanation than has been found before (line~13). 
If that is the case, we block the seed and all of its subsets by appending a
blocking clause to $M$ (line~14, where $\phi_i$ is a variable in the CNF formula).
Otherwise, we query the oracle for satisfiability with the new seed (line~15). 
If the seed is satisfiable, we climb up in the lattice until we find a MSS
(line~16), and block all of its subsets by appending a blocking clause to $M$
(line~17). We then compute the cardinality of the seed that yields an
explanation that is less verbose than any of the previously ones encountered 
(line~19), and block all smaller ones (line~20).

Otherwise, we climb down in the lattice until we find a MUS, and block all of
its supersets (lines~22--23). This process is then repeated until the SAT solver 
concludes that $M$ is unsatisfiable, in which case all MSSes have been 
considered. Finally, the algorithm returns a set of variable indices present 
in an explanation that is minimum with respect to $g$ (line~26).

\begin{proposition}[Correctness]
\label{prop:minimum}
Let $F$ be a tree ensemble, $p: f_{bin}(c_1, \ldots, c_n; F) \mapsto d$
a prediction, and $W = (w_1, \ldots, w_n)$ an $n$-tuple of positive weights. The set 
returned by Algorithm~\ref{algo:minimum-marco} contains the indices of a minimum
explanation for $p$ with respect to the cost function $g(I; W) = \sum_{i \in I}w_i$.
\end{proposition}

\section{Experimental Study}
\label{sec:comp-study}
In this section, we first evaluate the algorithms proposed in 
Section~\ref{sec:algos} in terms of runtime performance in different 
classification problems from related work\cite{Ignatiev22} (Section~\ref{sec:perf}).
We then take a closer look at the characteristics of explanations computed 
during our performance measurements (Section~\ref{sec:characteristics}). 
In particular, we investigate how many minimal and minimum explanations
there are for a given model and prediction, and how many of the variables in
the model are present in a given explanations. Finally, we revisit our running
example and provide concrete examples on how explanations can look like for
Android malware classifications (Section~\ref{sec:concrete-example}).

\subsection{Performance Evaluation}
\label{sec:perf}
To assess the performance of our proposed algorithms, we reuse 21 gradient 
boosting machines trained by Ignatiev~et~al.\cite{Ignatiev22} using XGBoost 
(version~1.2.1), configured for 50 trees with a max tree depth between 3 and 4. 
The number of input variables varies between 7--60, and the number of classes 
between 2--11. There are 200 predictions per dataset in need of explanation, 
with a few exceptions due to limited amount of data. 

\subsubsection{Minimal Explanations}
\label{sec:minimal-exp-eval}
To assess the scalability of Algorithm~\ref{algo:minimal}, we first implement
it as an extension to VoTE (version~0.3.0), which operates on elements from the
interval domain, as defined by Table~\ref{tbl:intdom}. We then revisit 
experiments conducted by Ignatiev~et~al.\cite{Ignatiev22} where a minimal explanation 
is computed for a specific prediction using two different approaches 
implemented in a tool named XReason: one using an SMT solver, and another
using a MaxSAT solver. Specifically, the first approach uses the SMT solver
Z3\cite{Z3} provided via pySMT\cite{pySMT}, and the second one 
uses an incremental MaxSAT solver based on RC2\cite{RC2} together with
the SAT solver Glucose3\cite{Glucose3}, both also provided via
pySAT. Since all of the experiments from the original study
are fully automated and published online, we are able to create a couple of
additions to the benchmarking scripts to also include our implementation of 
Algorithm~\ref{algo:minimal}.

To rerun the experiments, we invoke the scripts with the same arguments as
in the original study on a workstation equipped with an AMD Ryzen~7 3700X 
CPU and 32\,GiB of RAM, operated by Ubuntu~22.04. All experiments are executed
sequentially one after another, utilizing at most a single CPU core at any 
given point in time.

Table~\ref{tbl:minimal-cmp1} lists characteristics of the datasets associated
with the experiments, together with the elapsed runtime (in seconds) for the 
evaluated approaches (SMT, MaxSAT, and Abstract interpretation). The 
characteristics include the name of the dataset (name), the number of inputs 
that the trees accept (inputs), the number of classes in the output domain 
(classes), and the number of predictions that need an explanation (samples). 
Source code to reproduce these results are available online\footnote{Published
online, should the paper be accepted.}, together with logs from the runs that 
are the basis for the results presented here.

\begin{table}[ht]
  \centering
  \caption{Characteristics of datasets together with the elapsed time (in 
           seconds) per minimal explanation when using an SMT solver, a MaxSAT 
           solver, and abstract interpretation to explain predictions made by 
           ensembles of 50 trees with a max depth between 3--4.}
  \begin{tabular}{|l|r|r|r|r|r|r|r|r|r|r|r|r|}
    \hline
    \multicolumn{4}{|c|}{\textbf{Dataset}}                &
    \multicolumn{3}{c|}{\textbf{SMT}}                     &
    \multicolumn{3}{c|}{\textbf{MaxSAT}}                  &
    \multicolumn{3}{c|}{\textbf{Abstract interp.}} \\
    \hline
    \textbf{name} & \textbf{inputs} & \textbf{classes} & \textbf{samples} & 
    \textbf{min} & \textbf{avg} & \textbf{max} & 
    \textbf{min} & \textbf{avg} & \textbf{max} & 
    \textbf{min} & \textbf{avg} & \textbf{max} \\
    \hline
         ann-thyroid & 21 & 3  & 200 & 0.02 & 0.06 & 2.08 & 0.03 & 0.06 & 0.30 & 0.00 & 0.00 & 0.01 \\
        appendicitis & 7  & 2  & 106 & 0.01 & 0.03 & 0.09 & 0.02 & 0.02 & 0.04 & 0.00 & 0.00 & 0.00 \\
      biodegradation & 41 & 2  & 200 & 0.20 & 2.12 & 48.21 & 0.60 & 1.55 & 2.95 & 0.00 & 0.01 & 0.08 \\
             divorce & 54 & 2  & 150 & 0.03 & 0.04 & 0.06 & 0.00 & 0.01 & 0.17 & 0.00 & 0.00 & 0.00 \\
               ecoli & 7  & 5  & 200 & 0.08 & 0.45 & 4.63 & 0.20 & 0.53 & 0.88 & 0.00 & 0.00 & 0.01 \\
              glass2 & 9  & 2  & 162 & 0.03 & 0.09 & 0.28 & 0.08 & 0.14 & 0.21 & 0.00 & 0.00 & 0.00 \\
          ionosphere & 34 & 2  & 200 & 0.14 & 0.53 & 5.53 & 0.22 & 0.39 & 0.61 & 0.00 & 0.00 & 0.00 \\
           pendigits & 16 & 10 & 110 & 3.63 & 280.71 & 4772.60 & 5.17 & 10.61 & 18.44 & 0.02 & 0.08 & 0.52 \\
           promoters & 58 & 2  & 106 & 0.03 & 0.03 & 0.04 & 0.00 & 0.00 & 0.00 & 0.00 & 0.00 & 0.00 \\
        segmentation & 19 & 7  & 200 & 0.15 & 1.09 & 7.54 & 0.09 & 0.40 & 1.04 & 0.00 & 0.01 & 0.02 \\
             shuttle & 9  & 7  & 200 & 0.07 & 0.24 & 2.82 & 0.07 & 0.18 & 0.27 & 0.00 & 0.00 & 0.01 \\
               sonar & 60 & 2  & 200 & 0.17 & 0.53 & 7.59 & 0.27 & 0.42 & 0.65 & 0.00 & 0.00 & 0.02 \\
            spambase & 57 & 2  & 200 & 0.29 & 2.87 & 57.19 & 1.84 & 6.20 & 20.16 & 0.00 & 0.03 & 1.00 \\
    \textbf{texture} & 40 & 11 & 200  & 5.87 & 60.66 & 604.45 & 6.22 & 15.52 & \textbf{27.46} & 0.02 & 0.13 & \textbf{1.39} \\
            threeOf9 & 9  & 2  & 200  & 0.00 & 0.01 & 0.01 & 0.00 & 0.00 & 0.00 & 0.00 & 0.00 & 0.00 \\
             twonorm & 20 & 2  & 200 & 0.11 & 0.57 & 5.21 & 0.56 & 0.95 & 1.37 & 0.00 & 0.00 & 0.01 \\
               vowel & 13 & 11 & 200 & 1.79 & 30.17 & 233.95 & 4.74 & 7.95 & 11.80 & 0.00 & 0.02 & 0.06 \\
                wdbc & 30 & 2  & 200 & 0.08 & 0.23 & 0.55 & 0.15 & 0.26 & 0.34 & 0.00 & 0.00 & 0.00 \\
    wine-recognition & 13 & 3  & 178 & 0.02 & 0.06 & 0.19 & 0.02 & 0.05 & 0.09 & 0.00 & 0.00 & 0.00 \\
                wpbc & 33 & 2  & 194 & 0.17 & 0.80 & 4.80 & 0.29 & 1.10 & 2.77 & 0.00 & 0.00 & 0.02 \\
                 zoo & 16 & 7  & 59  & 0.06 & 0.20 & 0.60 & 0.01 & 0.01 & 0.05 & 0.00 & 0.00 & 0.00 \\
    \hline
  \end{tabular}
  \label{tbl:minimal-cmp1}
\end{table}

As observed in the original experiments, we confirm that the SMT approach is 
significantly slower than the MaxSAT approach. Furthermore, we see that the 
approach based on abstract interpretation computes minimal explanations in less
time than the MaxSAT approach. In the most time-consuming scenario (texture), 
the MaxSAT approach took 27.46 seconds to compute a single explanation, while
abstraction interpretation only took 1.36 seconds, amounting to a speedup factor
of 20. In total, the MaxSAT approach needed about 2.6 hours to compute a minimal
explanation for all the predictions, while the approach based on abstract
interpretation only needed 58 seconds, amounting to an overall speedup factor
of 157.

This significant speedup is surprising. To validate the correctness of the
implementation of Algorithm~\ref{algo:minimal}, we compare the explanations
themselves. The MaxSAT approach often provides different explanations compared
to the other two approaches, suggesting that the MaxSAT approach explores 
elements in the power set lattice of the constraint system in a different order
then the other two. When comparing explanations computed by the SMT
approach and abstract interpretation, we obtain the same results, with two
exceptions. To check the validity of the two explanations computed by the
SMT approach in those exceptions, we can query the oracle based on abstract 
interpretation, and record the refined abstract input value $\absvec{X}$ 
whenever the property checker (line~2 in Algorithm~\ref{algo:VoTE}) yields 
the outcome $Fail$, and extract concrete counterexamples of predictions where
the explanation is not valid by applying the concretization function $\gamma$ 
to $\absvec{X}$. 
As it turns out, invoking the original prediction function in XGBoost with these
counterexamples yields different outputs than the prediction subject to 
explanation. Consequently, the SMT approach implemented in XReason compute
two explanations that are not valid, hence that tool is not sound. After 
inspecting the source code of XReason, we believe these exceptions could
be caused by the fact the implemented SMT approach encodes variables in tree
ensembles as reals rather than floats, which may lead to incorrect query 
responses\cite{Brain15}. 

\begin{figure}[ht]
  \center
   \includegraphics[scale=0.8]{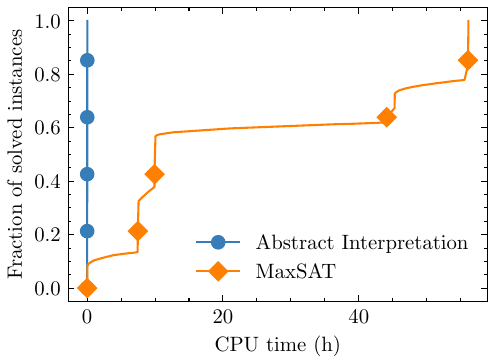}
   \caption{Progression of the MaxSAT and Abstract Interpretation approach when
            computing minimal explanations, where the max depth of trees are in
            the range $[6..8]$.}
    \label{fig:minimal-burnup}
\end{figure}

To further explore the scalability of Algorithm~\ref{algo:minimal}, we retrain
the gradient boosting machines with a max tree depth in the range $[6..8]$,
and rerun the experiments. Since the SMT approach is significantly slower than
the other two, we make additional changes to the source code of the automated
experiments to omit the SMT approach from further evaluation. We execute the
experiments, and note that the MaxSAT approach ran for a total of 56 hours, 
while the abstract interpretation approach only ran for 81 seconds, amounting
to a speedup factor of approximately 2,500. A burnup chart of the two
approaches is presented in Figure~\ref{fig:minimal-burnup}, which illustrates
the fraction between the number of computed minimal explanations and the number
of predictions subject to explanation at a given point in time during the
experiments.

Finally, we can gain some insights into the scalability of the two approaches by
summarizing their overall elapsed runtimes when computing minimal explanations
with different max tree depths, as illustrated by Table~\ref{tbl:minimal-slowdown}.

\begin{table}[ht]
  \centering
  \caption{Summary of elapsed runtimes when computing minimal explanations
  for predictions made by ensembles with different oracles and tree depths.}
  \begin{tabular}{c|r|r}
    \hline
    \textbf{Oracle approach} & 
    \textbf{Max tree depth} &
    \textbf{Elapsed runtime} \\
    \hline
    Abstract                & $[3..4]$ & 58\,s \\
    Interpretation          & $[6..8]$ & 81\,s \\
    \hdashline                            
    \multirow{2}{*}{MaxSAT} & $[3..4]$ & 2.6\,h \\
                            & $[6..8]$ & 56\,h \\
    \hline
  \end{tabular}
  \label{tbl:minimal-slowdown}
\end{table}
Specifically, we can see that doubling the tree depths yields a slowdown factor
of 21.5 for the MaxSAT approach, and only a factor of 1.4 for the approach based
on abstract interpretation.

\subsubsection{Minimum Explanations}
To assess the runtime performance of Algorithm~\ref{algo:minimum-marco}, we 
again reuse the datasets and models from Ignatiev~et~al.\cite{Ignatiev22}, but now aim to
compute explanations that are minimum with respect to the cost function 
$g(I; W) = \sum_{i \in I}w_i$. Due to a lack of domain knowledge in the many 
different datasets, we use the default weighs $W = \{1\}^n$, hence $g(I; W) = |I|$.
For seed generation with the m-MARCO algorithm, we use the SAT solver Gluecard3 
from PySAT\cite{pySAT}, which is a modification of Glucose3\cite{Glucose3} 
with added support for native cardinality constraints as published by
Liffiton~and~Maglalang\cite{Liffiton12}.

As baselines for comparison, we implement two alternative algorithms with 
formal descriptions in Appendix~\ref{sec:minimum-alt}. The first algorithm is 
based on a branch-and-bound approach (BB) realized in some SMT
solvers\cite{Dillig12}, while the second one is based on minimum
hitting sets (MHS)\cite{Ignatiev15}. For the MHS approach, we use 
the incremental MaxSAT solver RC2\cite{RC2} together with the SAT solver 
Glucose3\cite{Glucose3}, both provided via pySAT\cite{pySAT}.

All three algorithms are invoked by a program that utilize a single CPU core at
a time, configured to abort the minimization of a single explanation after one
hour. The experiments run on a compute cluster operated by Rocky~Linux~9.3, 
where each compute node is equipped with an Intel Xeon Gold 6130 processor. 
For each of the three approaches, Table~\ref{tbl:minimum} lists the total 
elapsed runtime (in hours), how much of that time was spent in the oracle (querytime), 
the total number of queries made to the oracle, and the percentage of solved 
instances. All experiments that led to unsolved instances were terminated by
the 60 minute timeout, see Appendix~\ref{sec:detailed-results} for details on
which models this occurred with, and on how many instances.
\begin{table}[ht]
  \centering
  \caption{Elapsed time, number of oracle queries, and percentage of solved
           instances for the different approaches that compute minimum
           explanations.}
  \begin{tabular}{l|r|r|r|r}
    \hline
    \textbf{Approach} & 
    \textbf{Runtime (h)} &
    \textbf{Querytime (h)} &
    \textbf{\# queries} &
    \textbf{Solved instances (\%)} \\

    \hline
      BB      &  1,822.0 &   1,798.7 &  $56  \cdot 10^9$ &  54.01 \\
      MHS     &     50.8 &      49.3 &  $354 \cdot 10^6$ &  99.79 \\
      m-MARCO &     19.1 &      18.4 &  $131 \cdot 10^6$ &  99.97 \\
    \hline
  \end{tabular}
  \label{tbl:minimum}
\end{table}

We see that the BB approach is an order of magnitude slower than the other two, 
solving about 54\% of the problem instances in 1,822 hours. We can also see that
the m-MARCO approach is 2.7 times faster than MHS, and manages to solve 99.97\% 
of the problems in 19 hours compared to 99.79\% in about 51 hours for MHS. 
Furthermore, we see that there is a strong correlation between the elapsed 
runtime and number of queries made to the oracle (large number of queries 
yields a large runtime), and that the MHS and m-MARCO approach only spend a
small portion of its runtime outside the oracle. Consequently, there is limited
performance to gain by optimising the underlying SAT or MaxSAT solver. 

\begin{figure}[ht]
  \center
  \includegraphics[scale=1]{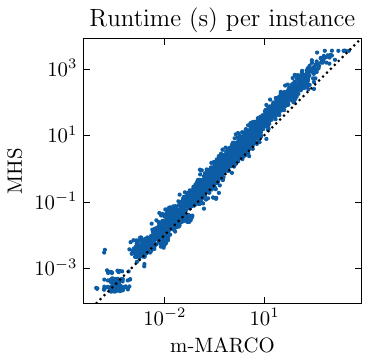}
  \includegraphics[scale=1]{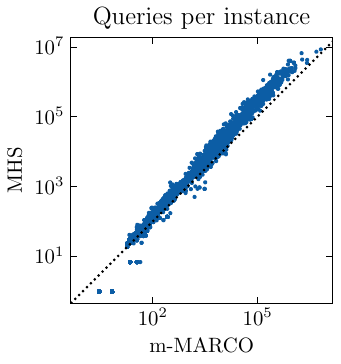}
   \caption{Elapsed runtime (to the left) and number of queries made to the 
            oracle (to the right) per instance when using the MHS and m-MARCO
            approach.}
    \label{fig:minimum-scatter}
\end{figure}

To gain further insights on how the m-MARCO approach perform compared to the
MHS approach, we compare the elapsed runtime and number of oracle queries the 
two approaches made for each problem instance, as illustrated by 
Figure~\ref{fig:minimum-scatter}. 
Here, we can see that m-MARCO is typically both faster and makes fewer queries
to the oracle than MHS on most of the instances, especially for the more time
consuming experiments.

\begin{figure}[ht]
  \center
  \includegraphics[scale=1]{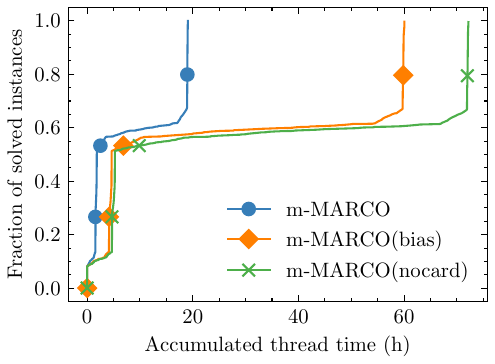}
   \caption{The progression of different variants of the m-MARCO approach
            when computing minimum explanations.}
    \label{fig:minimum-burnup}
\end{figure}

In the pursuit for improved runtime performance of m-MARCO, we explore
strategies that aim to reduce the number of queries made to the oracle when
shrinking seeds. In particular, Liffiton~et~al. propose MARCO+\cite{Liffiton16}, which
is able to systematically work its way towards the frontier between MUSes and
MSSes from the top of the power set lattice towards the bottom. In doing so, 
we know that whenever a seed is satisfiable, it is also a MSS. Consequently, 
we do not have to invoke the grow procedure at all (line~16 in 
Algorithm~\ref{algo:minimum-marco}). We implement this strategy in the SAT-based
seed generator, remove the calls to the grow procedure, and refer to this
variant as m-MARCO(bias). To further understand the impact the cardinality 
constraints has on performance, we also implement a variant where those 
constraints are omitted (line~20 in Algorithm~\ref{algo:minimum-marco}),
calling it m-MARCO(nocard). A burnup chart of the different variants of m-MARCO
is presented in Figure~\ref{fig:minimum-burnup}, which illustrates the 
fraction between the number of computed minimum explanations and the number
of predictions subject to explanation after a certain amount of accumulated
CPU thread time.

Here, we can clearly see that the use of cardinality constraints greatly
improves the runtime performance of m-MARCO, overall by a factor of 3.8.
Furthermore, we can see that m-MARCO also outperforms m-MARCO(bias), overall 
by a factor of 3.1. This suggests that in our experiments, the main bottleneck 
does not reside directly in the MARCO algorithm, but rather that the order in 
which seeds are generated is more relevant to the runtime performance. 
Being able to identify which order to generate seeds in for a given prediction
model is an interesting problem that deserves some attention, but is out of 
scope for this work.

\subsubsection{Enumerating All Minimal Explanations}
\label{sec:runtime-enumerate}
\begin{figure}[ht]
  \center
   \includegraphics[scale=1]{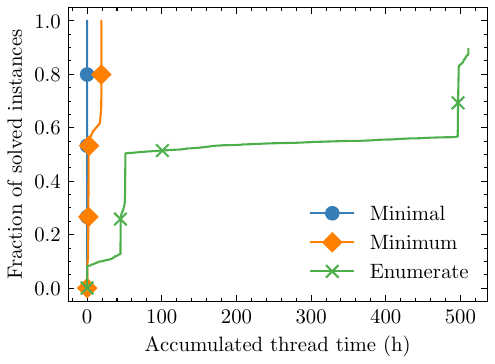}
   \caption{Progress made while computing minimal explanations
           (Algorithm~\ref{algo:minimal}), minimum explanations
           (Algorithm~\ref{algo:minimum-marco}), and when enumerating
           all minimal explanations (Algorithm~\ref{algo:marco}).}
  \label{fig:enumerate-burnup}
\end{figure}

We also implement the standard MARCO approach (Algorithm~\ref{algo:marco}) that 
enumerates all minimal explanations, and execute the same experiments on the
same compute cluster as before. Figure~\ref{fig:enumerate-burnup} illustrates
a burnup chart in the same format as before, but now comparing
Algorithms~\ref{algo:marco}~and~\ref{algo:minimum-marco}. We also include
Algorithm~\ref{algo:minimal} in the same figure to illustrate the
trade-off in terms of computational cost one is faced with when choosing
between computing a minimal explanation or a minimum one. 

We can clearly see that computing a minimal explanation using
Algorithm~\ref{algo:minimal} is significantly faster than computing a minimum
explanation using Algorithm~\ref{algo:marco}, which is expected given their 
theoretical computational complexity being $\text{D}^\text{P}$-complete and
$\Sigma_2^p$-complete, respectively. We also see that computing a minimum 
explanation with Algorithm~\ref{algo:minimum-marco} is faster than enumerating
all minimal explanations in practice. Overall, we managed to enumerate all 
minimal explanations for about 89\% of the predictions, with an accumulated 
CPU thread runtime of 511 hours, which is about 27 times as long compared to 
Algorithm~\ref{algo:minimum-marco} that computes a single minimum explanation
per prediction.

\subsection{Characteristics of Explanations}
\label{sec:characteristics}
In this section, we take a closer look at the explanations we enumerated in
Section~\ref{sec:runtime-enumerate}, using the standard MARCO algorithm 
(Algorithm~\ref{algo:marco}) paired with our valid explanation oracle 
(Algorithm~\ref{algo:oracle}). Note that we are now just interested in the 
actual explanations, hence the choice of oracle is irrelevant for this study. 
Any oracle that satisfies Definition~\ref{def:valid-exp-oracle} (valid 
explanation oracle) is adequate given enough runtime, which includes all 
approaches evaluated in Section~\ref{sec:minimal-exp-eval}. First, we 
investigate the number of (minimal and minimum) explanations that were 
enumerated for each trained model, and then the number of variables that are
present in each computed explanation.
\subsubsection{Number of Explanations}
Table~\ref{tbl:counts} lists the number of minimal and minimum explanations
that were enumerated in Section~\ref{sec:runtime-enumerate} using the oracle
based on VoTE (Algorithm~\ref{algo:oracle}), together with the total number
of input variables each model accepts, and how many variables that are 
referenced during prediction (i.e., $V_f$, as defined by 
Definition~\ref{def:var-ref-tree}).
\begin{table}[ht]
  \centering
  \caption{Characteristics of input variables of models trained on different 
           datasets, together with the number of minimal and minimum 
           explanations enumerated using the standard MARCO approach 
           (Algorithm~\ref{algo:marco} combined with Algorithm~\ref{algo:oracle}).}
  \begin{tabular}{|r|r|r|r|r|r|r|r|r|}
    \hline
    \multicolumn{1}{|r|}{\textbf{Name of}} &
    \multicolumn{2}{|r|}{\textbf{Input variables}} &
    \multicolumn{3}{c|}{\textbf{Minimal explanations}}  &
    \multicolumn{3}{c|}{\textbf{Minimum explanations}}   \\
    \cline{2-9} 
    \textbf{dataset} & \textbf{referenced} & \textbf{total} & 
    \textbf{min} & \textbf{avg} & \textbf{max} & 
    \textbf{min} & \textbf{avg} & \textbf{max} \\   
    \hline 

     ann-thyroid &        11 &    21 &    1 &        1.5 &        7 &  1 &   1.0  &     2 \\
    appendicitis &         7 &     7 &    1 &        7.6 &       17 &  1 &   3.3  &    11 \\
  biodegradation &        34 &    41 &    1 &    20,594.9 &   370,760 &  1 &  26.9  &   544 \\
         divorce &        14 &    54 &    7 &      126.8 &      177 &  1 &   3.9  &    19 \\
           ecoli &         6 &     7 &    1 &        2.4 &        6 &  1 &   2.1  &     6 \\
          glass2 &         8 &     9 &    1 &        5.9 &       16 &  1 &   2.7  &    10 \\
      ionosphere &        31 &    34 &   17 &    33,003.4 &   382,099 &  1 &  19.1  &   111 \\
       pendigits &        16 &    16 &    6 &      111.0 &      263 &  1 &   7.0  &    44 \\
       \textbf{promoters} &         \textbf{1} &    \textbf{58} &    1 &        1.0 &        1 &  1 &   1.0  &     1 \\
    segmentation &        17 &    19 &    3 &       40.0 &      229 &  1 &   5.4  &    38 \\
         shuttle &         9 &     9 &    1 &        2.6 &        8 &  1 &   1.7  &     5 \\
           \textbf{sonar} &        \textbf{50} &    60 &  191 &   168,215.8 &   \textbf{576,834} &  1 &  56.9  &   \textbf{609} \\
        spambase &        49 &    57 &    7 &    36,915.2 &   202,322 &  1 &  20.7  &   138 \\
         texture &        37 &    40 &  987 &    13,284.4 &    25,365 &  1 &  11.3  &   107 \\
        threeOf9 &         3 &     9 &    1 &        1.0 &        1 &  1 &   1.0  &     1 \\
         twonorm &        20 &    20 &    1 &     2,530.6 &    17,682 &  1 &  21.7  &   607 \\
           vowel &        13 &    13 &    1 &       11.7 &       40 &  1 &   3.8  &    19 \\
            wdbc &        27 &    30 &    3 &     6,971.3 &    39,889 &  1 &   9.0  &    70 \\
wine-recognition &        12 &    13 &    1 &        5.1 &       14 &  1 &   2.4  &     6 \\
            wpbc &        32 &    33 &   43 &    32,791.4 &   297,748 &  1 &  16.1  &   267 \\
             zoo &        13 &    16 &    1 &        4.1 &       10 &  1 &   2.0  &     6 \\
    \hline
  \end{tabular}
  \label{tbl:counts}
\end{table}

We can clearly see that the number of explanations varies greatly, both between 
different models and between different predictions made by the same model.
We also see that the number of referenced variables seems to have a large 
impact on the average (avg) and maximal (max) number of explanations computed
for a particular prediction. Furthermore, we see that there are 
significantly fewer minimum than minimal explanations, often by several 
orders of magnitude. In the most extreme case (the sonar dataset), there
are 50 referenced variables, up to 576,834 minimal explanations for a 
single prediction, and up to 609 minimum explanations.
Another insight is that there are several models in which very few of the 
possible variables are actually used in predictions (i.e. referenced by 
the model). The extreme case is the promoters model, which uses only a single 
input variable out of the 58 possible ones.
\subsubsection{Verbosity of Explanations}
\begin{figure}[ht]
  \center
   \includegraphics[scale=0.85]{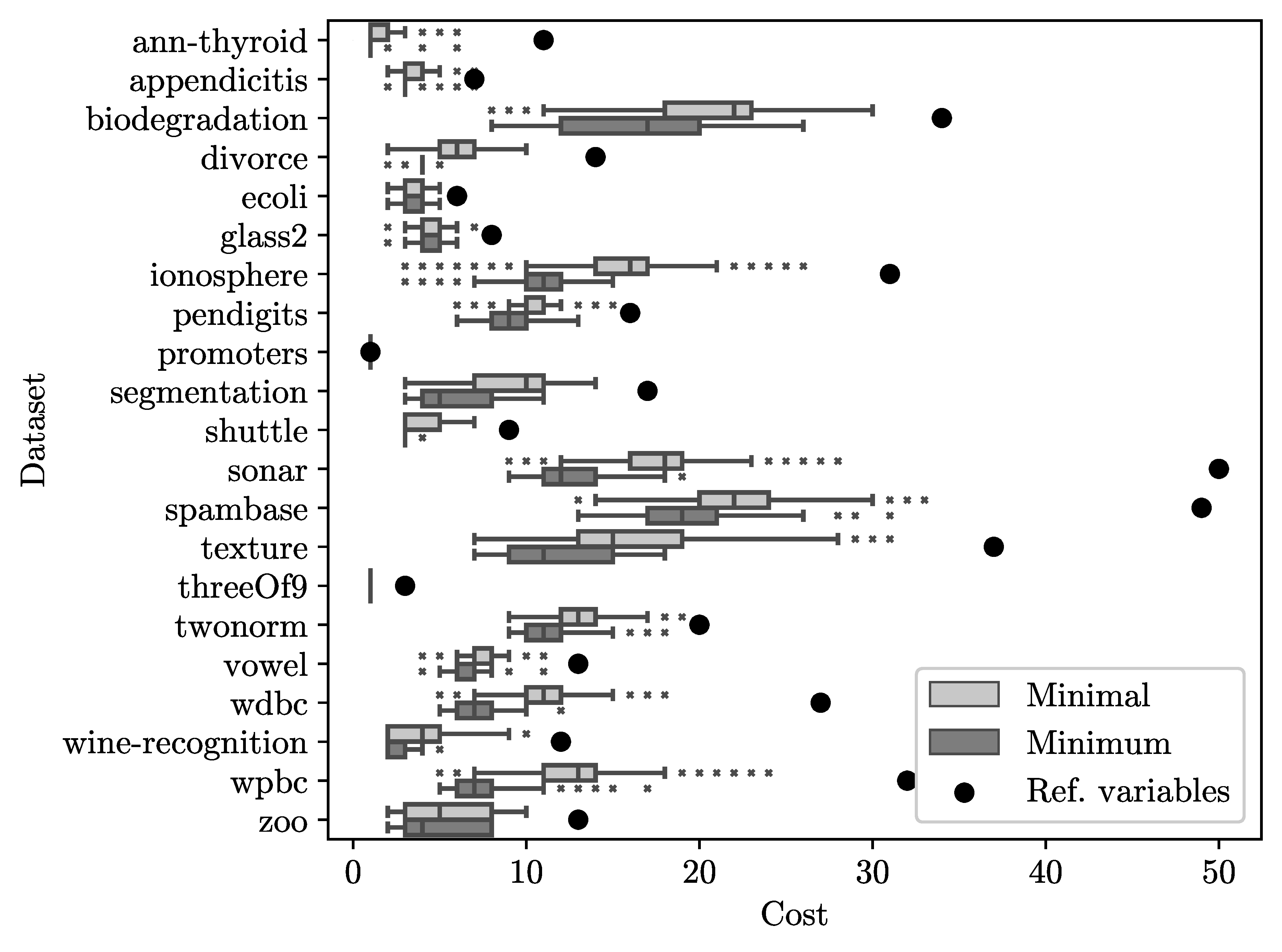}
   \caption{A boxplot illustrating costs of different explanations.}
  \label{fig:costs}
\end{figure}

Figure~\ref{fig:costs} illustrates a boxplot with costs associated with
the verbosity of minimal and minimum explanations for a given model, computed
using the cost function $g(I) = |I|$. The figure also illustrates 
the number of referenced variables in each model with filled circular data 
points, i.e., $g(V_f)$.

Again, we observe that models with few referenced variables stand out.
The threeOf9 model refers to three variables, but all explanations enumerated 
in our experiments contain a single variable. The promoters model
only references a single input variable during prediction, hence all 
explanations for those prediction cost the same. These are models that provide
fewer insights in terms of minimization. 

The rest of the chart illustrates that significant insights regarding the
characteristics of a model can be obtained by performing computations of 
minimal and minimum explanations. Recall that a circular data point reveals
the number of variables in a trivial (verbose) explanation from a model. 
Then the gap between the circles and the median values for each model reveals
that minimization of explanations pays off in terms of verbosity, but the 
nature of the model dictates the size of the gap.

By cross-referencing with Table~\ref{tbl:counts}, we can also see that models 
that yield large numbers of minimal explanations typically have a large spread
of the cost associated with them, e.g., the biodegradation and the sonar model.
In these cases, we can also see a significant difference in costs when 
comparing minimal and minimum explanations.

\subsection{Android Malware Classification}
\label{sec:concrete-example}
To illustrate how explanations may look like in practice, we revisit the Android
malware classification problem introduced in Section~\ref{sec:cls-example}.
Here, we use XGBoost to train 100 trees with depth 10 under monotone constraints, 
i.e., ensuring that potential adversaries that add more permissions to a malware 
does not cause the classifier to change its prediction to benign. We then use
Algorithm~\ref{algo:minimal} to compute a minimal explanation for an application
that requests all of the permissions, yielding an explanation with the following
five permissions:
\begin{itemize}
  \setlength\itemsep{0em}
  \item load plugins in browser,
  \item write browser history and bookmarks,
  \item install shortcuts,
  \item read device settings,
  \item uninstall shortcuts.
\end{itemize}
Finally, we use Algorithm~\ref{algo:minimum-marco} to compute a minimum 
explanation for that same application, yielding an explanation with only two 
permissions:
\begin{itemize}
  \setlength\itemsep{0em}
  \item can send SMS,
  \item keep Wi-Fi awake.
\end{itemize}
\section{Conclusions}
\label{sec:conclusions}
When predictions made by machine learning models are used for decision support 
in critical systems operated by humans, explanations can help the operators 
to make informed interventions of actions when their explanations appear unjustified. 
In this context, the explanations must be correct, and preferably without 
redundant information. The goal of this work has been to provide such 
explanations efficiently. To this end, we have formalized a sound and complete
reasoning engine based on abstract interpretation, tailored specifically for 
determining the validity of explanations for predictions made by tree ensembles.
The reasoning engine is composed of an oracle that can determine the validity 
of explanations, which is paired with state-of-art algorithms (deletion filter 
and MARCO) to compute and enumerate minimal explanations. Furthermore, we have
presented a novel adaptation of the MARCO algorithm called m-MARCO to compute
an explanation that is minimum with respect to a cost function.

We demonstrate runtime performance speedups by several orders of magnitude 
compared to current state-of-the-art. We also show that, in the presence of a 
highly efficient oracle, it is possible to compute a minimum explanation for 
several non-trivial tree ensembles. Through enumeration of all minimal 
explanations, we were able to examine the nature of the obtained explanations. 
For example, we show that there are typically a large number of minimal 
explanations to choose from, sometimes exceeding 500,000 for a single 
prediction. In these cases, we learned that minimum explanations are typically 
significantly less verbose than minimal explanations, but also more 
time-consuming to compute.

We have also learned that some fundamental algorithms that off-the-shelf
constraint solvers rely on can be tailored to narrower classes of problems, 
in favor of significant increases in runtime performance, thus facilitating
computer-aided problem-solving that was previously intractable. Our analysis 
of models in multiple applications from the literature shows that the models 
in themselves often have superfluous variables, which would appear in 
explanations without systematic minimization. While some works may aim to 
improve models by removing unused features, a gap still remains between the 
features actually used in a model and the least verbose explanations for a 
given prediction. This gap motivates striving for correct and compact 
explanations using approaches like ours. 

As machine learning models mature into well-established and standardized
architectures, and their applications become more critical to society, we 
believe there is a lot to gain by continuing this line of research to include
even more types of models and prediction functions like random forests, support
vector machines and neural networks, where generic constraint solvers are
likely to serve as essential oracles when validating implementations of
model-specific ones.


\bibliography{main}

\begin{thebibliography}{10}
\providecommand \doibase [0]{http://dx.doi.org/}%

\bibitem{Hadji21}
Hadji~Misheva B, Jaggi D, Posth JA, Gramespacher T, Osterrieder J.
  Audience-Dependent Explanations for {AI}-Based Risk Management Tools: A
  Survey. {\it Frontiers in Artificial Intelligence.} 2021\string;4.

\bibitem{Ras22}
Ras G, Xie N, Gerven vM, Doran D. Explainable Deep Learning: A Field Guide for
  the Uninitiated. {\it Journal of Artificial Intelligence Research.}
  2022\string;73\string:329--397.

\bibitem{Kaur22}
Kaur D, Uslu S, Rittichier KJ, Durresi A. Trustworthy Artificial Intelligence:
  A Review. {\it ACM Computing Surveys (CSUR).} 2022\string;55(2)\string:1--38.

\bibitem{Chen16}
Chen T, Guestrin C. {XGB}oost: A Scalable Tree Boosting System. In: KDD '16.
  Proceedings of the 22nd ACM SIGKDD International Conference on Knowledge
  Discovery and Data Mining. Association for Computing Machinery 2016; New
  York, NY, USA\string:785–794.

\bibitem{Breiman01}
Breiman L. Random forests. {\it Machine learning.}
  2001\string;45(1)\string:5--32.

\bibitem{Friedman01}
Friedman JH. Greedy Function Approximation: A Gradient Boosting Machine. {\it
  The Annals of Statistics.} 2001\string;29(5)\string:1189--1232.

\bibitem{Arp14}
Arp D, Spreitzenbarth M, Hubner M, Gascon H, Rieck K, Siemens C. {{DREBIN}}:
  Effective and explainable detection of android malware in your pocket. In: .
  14. Network and distributed system security symposium.  2014\string:23--26.

\bibitem{Quine52}
Quine WV. The problem of simplifying truth functions. {\it The American
  mathematical monthly.} 1952\string;59(8)\string:521--531.

\bibitem{Marquis91}
Marquis P. Extending abduction from propositional to first-order logic. In:
  International Workshop on Fundamentals of Artificial Intelligence Research.
  1991\string:141--155.

\bibitem{Shih18}
Shih A, Choi A, Darwiche A. A symbolic approach to explaining Bayesian network
  classifiers. In: Proceedings of the 27th International Joint Conference on
  Artificial Intelligence.  2018\string:5103--5111.

\bibitem{Ignatiev19}
Ignatiev A, Narodytska N, Marques-Silva J. Abduction-based Explanations for
  Machine Learning Models. In: . 33. Proceedings of the AAAI Conference on
  Artificial Intelligence.  2019\string:1511--1519.

\bibitem{La21}
La~Malfa E, Zbrzezny A, Michelmore R, Paoletti N, Kwiatkowska M. On guaranteed
  optimal robust explanations for {NLP} models. In: 30th International Joint
  Conference on Artificial Intelligence.  2021\string:2658-2665.

\bibitem{Ignatiev22}
Ignatiev A, Izza Y, Stuckey PJ, Marques-Silva J. Using {MaxSAT} for Efficient
  Explanations of Tree Ensembles. In: Proceedings of the Thirty-Sixth {AAAI}
  Conference on Artificial Intelligence. {AAAI} Press 2022.

\bibitem{Audemard22}
Audemard G, Bellart S, Bounia L, Koriche F, Lagniez JM, Marquis P. Trading
  Complexity for Sparsity in Random Forest Explanations. In: Proceedings of the
  Thirty-Sixth {AAAI} Conference on Artificial Intelligence. {AAAI} Press 2022.

\bibitem{Darwiche20}
Darwiche A, Hirth A. On the Reasons Behind Decisions. In: In Proceedings of the
  24th European Conference on Artificial Intelligence (ECAI). {IOS} Press
  2020\string:712--720.

\bibitem{DARPA21}
Gunning D, Vorm E, Wang JY, Turek M. {DARPA}'s Explainable {AI} ({XAI})
  Program: A retrospective. {\it Applied AI Letters.} 2021\string;2(4).

\bibitem{Chinneck91}
Chinneck JW, Dravnieks EW. Locating minimal infeasible constraint sets in
  linear programs. {\it ORSA Journal on Computing.}
  1991\string;3(2)\string:157--168.

\bibitem{Marques21}
Marques-Silva J, Gerspacher T, Cooper MC, Ignatiev A, Narodytska N.
  Explanations for Monotonic Classifiers. In: International Conference on
  Machine Learning.  2021\string:7469--7479.

\bibitem{Ignatiev21}
Ignatiev A, Marques-Silva J. {SAT}-Based Rigorous Explanations for Decision
  Lists. In:  Li CM, Many{\`a} F. \kern-2pt, eds. {\it Theory and Applications
  of Satisfiability Testing -- SAT 2021}Theory and Applications of
  Satisfiability Testing -- SAT 2021. Springer International Publishing 2021;
  Cham\string:251--269.

\bibitem{Ignatiev20}
Ignatiev A, Narodytska N, Asher N, Marques-Silva J. From contrastive to
  abductive explanations and back again. In: International Conference of the
  Italian Association for Artificial Intelligence.  2020\string:335--355.

\bibitem{Liffiton13}
Liffiton MH, Malik A. Enumerating infeasibility: Finding multiple {MUS}es
  quickly. In: International Conference on Integration of Constraint
  Programming, Artificial Intelligence, and Operations Research.
  2013\string:160--175.

\bibitem{Previti13}
Previti A, Marques-Silva J. Partial {MUS} enumeration. In: Twenty-Seventh AAAI
  Conference on Artificial Intelligence.  2013.

\bibitem{Liffiton16}
Liffiton MH, Previti A, Malik A, Marques-Silva J. Fast, flexible {MUS}
  enumeration. {\it Constraints.} 2016\string;21(2)\string:223--250.

\bibitem{Cousot77}
Cousot P, Cousot R. Abstract Interpretation: A Unified Lattice Model for Static
  Analysis of Programs by Construction or Approximation of Fixpoints. In:
  Proceedings of the 4th ACM SIGACT-SIGPLAN symposium on Principles of
  Programming Languages (POPL). ACM Press, New York, NY 1977; Los Angeles,
  California\string:238--252.

\bibitem{Shih19}
Shih A, Choi A, Darwiche A. Compiling Bayesian network classifiers into
  decision graphs. In: . 33. Proceedings of the AAAI Conference on Artificial
  Intelligence.  2019\string:7966--7974.

\bibitem{Dillig12}
Dillig I, Dillig T, McMillan KL, Aiken A. Minimum satisfying assignments for
  {SMT}. In: International Conference on Computer Aided Verification.
  2012\string:394--409.

\bibitem{Ignatiev16}
Ignatiev A, Previti A, Marques-Silva J. On finding minimum satisfying
  assignments. In: International Conference on Principles and Practice of
  Constraint Programming.  2016\string:287--297.

\bibitem{Shwartz22}
Shwartz-Ziv R, Armon A. Tabular data: Deep learning is not all you need. {\it
  Information Fusion.} 2022\string;81\string:84--90.

\bibitem{Izza21}
Izza Y, Silva JM. On Explaining Random Forests with {SAT}. In: 30th
  International Joint Conference on Artificial Intelligence.  2021.

\bibitem{Boumazouza21}
Boumazouza R, Cheikh-Alili F, Mazure B, Tabia K. {\it ASTERYX: A Model-Agnostic
  SaT-BasEd AppRoach for SYmbolic and Score-Based
  EXplanations}\string:120–129; New York, NY, USA: Association for Computing
  Machinery .
\newblock 2021.

\bibitem{Tornblom19}
T{\"o}rnblom J, Nadjm-Tehrani S. An Abstraction-Refinement Approach to Formal
  Verification of Tree Ensembles. In: International Conference on Computer
  Safety, Reliability, and Security (SAFECOMP). Springer 2019\string:301--313.

\bibitem{Izza22}
Izza Y, Ignatiev A, Marques-Silva J. On Tackling Explanation Redundancy in
  Decision Trees. {\it Journal of Artificial Intelligence Research.}
  2022\string;75\string:61--321.

\bibitem{Ranzato20}
Ranzato F, Zanella M. Abstract Interpretation of Decision Tree Ensemble
  Classifiers. In: Proceedings of the Thirty-Fourth {AAAI} Conference on
  Artificial Intelligence. {AAAI} Press 2020\string:5478--5486.

\bibitem{Tornblom21}
T{\"o}rnblom J, Nadjm-Tehrani S. Scaling up memory-efficient formal
  verification tools for tree ensembles. {\it arXiv preprint.}
  2021\string;arXiv:2105.02595.

\bibitem{Z3}
De~Moura L, Bj{\o}rner N. Z3: An efficient {SMT} solver. In: International
  conference on Tools and Algorithms for the Construction and Analysis of
  Systems.  2008\string:337--340.

\bibitem{pySMT}
Gario M, Micheli A. {PySMT:} A solver-agnostic library for fast prototyping of
  {SMT}-based algorithms. In: SMT Workshop 2015.  2015.

\bibitem{RC2}
Ignatiev A, Morgado A, Marques-Silva J. {RC2}: An efficient {MaxSAT} solver.
  {\it Journal on Satisfiability, Boolean Modeling and Computation.}
  2019\string;11(1)\string:53--64.

\bibitem{Glucose3}
Audemard G, Lagniez JM, Simon L. Improving {G}lucose for incremental {SAT}
  solving with assumptions: Application to {MUS} extraction. In: International
  conference on theory and applications of satisfiability testing.
  2013\string:309--317.

\bibitem{Brain15}
Brain M, Tinelli C, R{\"u}mmer P, Wahl T. An automatable formal semantics for
  {IEEE}-754 floating-point arithmetic. In: 2015 IEEE 22nd Symposium on
  Computer Arithmetic.  2015\string:160--167.

\bibitem{pySAT}
Ignatiev A, Morgado A, Marques{-}Silva J. {PySAT:} A {Python} Toolkit for
  Prototyping with {SAT} Oracles. In: SAT.  2018\string:428--437

\bibitem{Liffiton12}
Liffiton MH, Maglalang JC. A Cardinality Solver: More Expressive Constraints
  for Free. In: Theory and Applications of Satisfiability Testing -- SAT 2012.
  Springer 2012\string:485--486.

\bibitem{Ignatiev15}
Ignatiev A, Previti A, Liffiton M, Marques-Silva J. Smallest {MUS} extraction
  with minimal hitting set dualization. In: International Conference on
  Principles and Practice of Constraint Programming.  2015\string:173--182.

\end{thebibliography}

\appendix

\section{Proofs}
\label{sec:proofs}

\subsection*{Lemma~\ref{lemma:tree}} 
\begin{proof}
Let $\bar{x} \in \D^n$ be an arbitrary value from the concrete input domain, 
$(X, y) \in T$ such that $\bar{x} \in X$, and thus $t(\bar{x}) = y$. 
According to Definition~\ref{def:conservative-transformation}, the transformer 
$\abs{t}$ is conservative iff $y \in \gamma(\abs{t}(\alpha_n(\{\bar{x}\}); T))$. 
By expanding $\abs{t}$ and using the fact that $t$ maps $\bar{x}$ to $y$, 
we obtain
\begin{equation*}
\begin{split}
\bar{y} \in\text{ }& \gamma(\abs{t}(\alpha_n(\{\bar{x}\}); T)) = \\
                     & \gamma(\alpha(\{y_i : (X_i, y_i) \in T, 
                                \alpha_n(\{\bar{x}\}) \sqcap \alpha_n(X_i) \ne \bot\})) = \\
                     & \gamma(\alpha(\{y\}))
\end{split}
\end{equation*} 
which holds when $\alpha$ and $\gamma$ form a Galois connection between the 
domains $\D$ and $\A$.
\end{proof}

\subsection*{Lemma~\ref{lemma:ensemble}} 
\begin{proof}
By using Lemma~\ref{lemma:tree} with the assumption on the abstract transformer
for the addition operator, all individual transformations performed by $\abs{f}$
are conservative, hence $\abs{f}$ is conservative.
\end{proof}

\subsection*{Lemma~\ref{lemma:sigmoid}} 
\begin{proof}
Let $Z \subseteq \D$ be an arbitrary subset of the concrete domain, $z \in Z$ 
an arbitrary element from that subset, 
$P: \{p_{\sigma}(\min(Z)), \ldots, p_{\sigma}(\max(Z))\}$, and
$p'_{\sigma}(z) = p_{\sigma}(z)(1 - p_{\sigma}(z))$ the derivative of $p_{\sigma}$.
Since $p'_{\sigma}$ is non-negative and thus the successive applications of 
$p_{\sigma}$ yield values that are monotonically increasing, we know that 
$p_{\sigma}(z) \in P$.
Furthermore, according to
Definition~\ref{def:conservative-transformation}, we know that the transformer
$\abs{p}_{\sigma}$ is conservative iff 
$p_{\sigma}(z) \in \gamma(\abs{p}_{\sigma}(\alpha(Z)))$.
By carefully applying $\alpha$ and $\gamma$ to appropriate terms in $P$, we obtain
\begin{equation*}
\begin{split}
p_{\sigma}(z)\in\text{ }& \{p_{\sigma}(\min(Z)), \ldots, p_{\sigma}(\max(Z))\} \subseteq \\
                  & \gamma(\alpha(\{p_{\sigma}(\min(Z)), \ldots, p_{\sigma}(\max(Z))\})) \subseteq \\
                  & \gamma(\alpha(\{p_{\sigma}(\min(\alpha(Z))), \ldots, p_{\sigma}(\max(\alpha(Z)))\})) = \\
                  & \gamma(\abs{p}_{\sigma}(\alpha(Z))),
\end{split}
\end{equation*} 
hence $\abs{p}_{\sigma}$ is conservative with respect to $p_{\sigma}$.
\end{proof}

\subsection*{Lemma~\ref{lemma:bin-classifier}} 
\begin{proof}
By using Lemma~\ref{lemma:ensemble}~and~\ref{lemma:sigmoid} together with the 
assumption on the abstract transformer for $>$, all individual transformations
performed by $\abs{f}_{bin}$ are conservative, hence $\abs{f}_{bin}$ is 
conservative.
\end{proof}

\subsection*{Lemma~\ref{lemma:precise}} 
\begin{proof}
In each recursive step, the algorithm picks a tree $T$, and computes abstract
input tuples $\absvec{x}_1, \ldots, \absvec{x}_k$ such that the tree transformer
$\abs{t}$ yields a precise output abstractions for that particular tree, i.e., 
$\forall i \in \{1, \ldots, k\}, |\gamma(\abs{t}(\absvec{x}_i; T))| = 1$.
When $F=R$, all trees in the ensemble have gone through a refinement step, hence
$\forall T \in F, \forall i \in \{1, \ldots, k\}, |\gamma(\abs{t}(\absvec{x}_i; T))| = 1$.
In these cases, the summation operator in $\abs{f}$ transforms abstract values
that capture a single point in the output space, hence the total sum is also 
an abstraction of a single point, i.e.
$\forall i \in \{1, \ldots, k\}, |\gamma(\abs{f}(\absvec{x}_i); F)| = 1$.
\end{proof}

\subsection*{Theorem~\ref{theo:sound}} 
\begin{proof}
Since $\absvec{x}$ captures all possible combinations of assignments to the
absent variables (as defined in Definition~\ref{def:abs-expl-oracle}), and
$\abs{f}_{bin}$ is conservative (according to Lemma~\ref{lemma:bin-classifier}), 
we know that when $d \not\in \gamma(\abs{f}_{bin}(\absvec{x}; F))$, the given
explanation is not valid, which is the only case where $v_{bin}$ returns $Fail$.
If all input points captured by $\abs{X}$ are mapped by $f_{bin}$ to $d$, 
i.e., $\{d\} = \gamma(\abs{f}_{bin}(\absvec{x}; F))$, the explanation is valid,
which is the only case where $v_{bin}$ returns $Pass$. Consequently, $v_{bin}$ 
is sound.
\end{proof}

\subsection*{Theorem~\ref{theo:correct}} 
\begin{proof}
By using Theorem~\ref{theo:sound}, we know that the algorithm is sound, i.e., 
the property checker $pc$ will only return $Pass$ when a given explanation is 
valid, and only return $Fail$ when it is not valid. By using 
Lemma~\ref{lemma:precise}, we know that VoTE will eventually invoke $pc$ with 
a sequence of abstract input tuples $\absvec{x}_p$ that form a partition of 
$\absvec{x}$ such that $|\gamma(\abs{y}_p)| = 1$, in which case
$pc(\absvec{x}_p, \abs{y}_p) \ne Unsure$. Consequently, the invocation of VoTE 
on line~11 will never return $Unsure$, hence the algorithm is complete.
\end{proof}

\subsection*{Proposition~\ref{prop:minimal}} 
\begin{proof}
Reducing the set of variables present in a valid explanation into a set that is 
minimal is equivalent to finding a MCS of a constraint system with elements that
encodes the absence of variables in an explanation. An MCS can be derived from
the complement of an MSS. Algorithm~\ref{algo:minimal} constructs an MSS by 
climbing towards the top of the power set lattice of the constraint system 
(which encodes the explanation where all variables are absent), starting at 
the bottom (which encoding the explanation where all variables are present), 
and queries a valid explanation oracle (Algorithm~\ref{algo:oracle}) for 
validity in each step. Since the oracle is both sound and complete, we are
guaranteed to find a minimal explanation.
\end{proof}

\subsection*{Proposition~\ref{prop:minimum}} 
\begin{proof}
Reducing the set of variables present in a valid explanation into a set that is 
minimum with respect to a cost function is equivalent to finding a minimum-cost 
MCS of a constraint system with elements that encode the absence of variables 
in an explanation. An MCS can be derived from the complement of an MSS, and
the MARCO algorithm\cite{Liffiton16} can be used to enumerate all MSSes.
In Algorithm~\ref{algo:minimum-marco}, we modify the standard MARCO algorithm
to search for a single MCS that is minimum with respect to the cost function 
$g$. Since we only need to find one MCS, we do not explore MCSes that cost 
more or equal to the cheapest MCS that we know of, which speeds up the search.
By combining the modified MARCO algorithm with the sound and complete valid 
explanation oracle formalized in Algorithm~\ref{algo:oracle}, we are thus 
guaranteed to find an explanation that is minimum with respect to $g$.
\end{proof}

\section{Multi-class classification}
\label{sec:multi-class}
For readability, previous sections only consider explanations for binary 
classifications. In this appendix, we extend our formalization to also include
explanations made by models trained in the \textit{one-vs-rest} classification 
paradigm. To this end, we first define the softmax function, which is used as
a post-processing function in favor of the sigmoid function as typically used
by binary classifiers.

\begin{definition}[Softmax Function]
\label{def:softmax}
The softmax function $p_s$ is a monotonic function that transforms a $j$-tuple
of real-valued inputs into an output tuple with elements in the range $[0, 1]$
such that those elements sum up to one, and is defined as
\begin{equation*}
  \label{eq:argmax}
  p_s(z_1, \ldots, z_j) = \frac{(e^{z_1}, \ldots, e^{z_j})}{\sum\limits_{i=1}^{j}z_i}.
\end{equation*}
\end{definition}
With the softmax function defined, we can define the tree-based multi-class 
classifier.
\begin{definition}[Tree-based Multi-class Classifier]
\label{def:multi-classifier}
Let $C=(F_1,\ldots,F_j$) be a collection of tree ensembles such that the $i$-th 
tree ensemble $F_i$ is trained to predict the probability that a given instance
belongs to the $i$-th class. A multi-class classifier $f_{cls}$ that 
discriminates between $j$ classes may then be defined as
\begin{equation*}
  \label{eq:multiclass-classifier}
  f_{cls}(\bar{x}; C) = 
    \min\Big(\argmax_{\{1, \ldots, j\}}\Big(p_s\big(f(\bar{x}; F_1), 
                                                      \ldots, 
                                                      f(\bar{x}; F_j)
                                                 \big)
                                        \Big)
    \Big)
\end{equation*}
\end{definition}
In principle, a multi-class classifier may predict that a given input belongs
to different classes with the exact same probability, and thus may cause the
argmax function to yield a set containing several classes. In these types of
situations, XGBoost return only the class with the smallest index, being 
indifferent to the selected choice of class.

\subsection*{Oracle for Multi-class Classifiers}
\label{sec:sup-algos}
For readability, Section~\ref{sec:algos} only considers explanations for 
binary classifications. In this section, we define an oracle that reasons about
collections of tree ensemble, which we used to minimize explanations of
predictions made by one-vs-rest classifiers in Section~\ref{sec:comp-study}. 
This oracle is formalized as 
Algorithm~\ref{algo:multi-oracle}, and accepts three inputs: a collection of 
$j$ tree ensembles $C$, a set of intervals $\absvec{x} \in \I^n$, and the 
predicted decision $d \in \{1, \ldots, j\}$.

\begin{algorithm}[ht]
  \caption{Check that a collection of tree ensembles $F_1, \ldots, F_j$ trained
           on a multi-class classification problem using the one vs. rest
           paradigm maps all points captured by the abstract input tuple 
           $\absvec{x} = (\abs{x}_1, \ldots, \abs{x}_n)$ to the label 
           $d \in \{1, \ldots, j\}$, where $j = |C|$ is the number of classes.}
  \label{algo:multi-oracle}
  \begin{algorithmic}[1]
    \Function{Multi\_Is\_Valid}{$C, \absvec{x}, d$}
      \Function{$pc_d$}{$\absvec{x}_d, \abs{y}_d$}
        \Function{$pc_i$}{$\absvec{x}_i, \abs{y}_i$}
          \Let{$D$}{$\gamma(\abs{y}_d > \abs{y}_i)$}
          \If{$|D| \ne 1$}
            \State \Return $Unsure$
          \ElsIf{$1 \in D$}
            \State \Return $Pass$
          \EndIf
        \State \Return $Fail$
        \EndFunction
        \ForEach{$i \in \{1 \ldots j\} \backslash \{d\}$}
          \Let{$o$}{\Call{VoTE}{$F_i, \absvec{x}_d, pc_i$}}
          \If{$o \ne Pass$}
            \State \Return $o$
          \EndIf
        \EndForEach
        \State \Return $Pass$
      \EndFunction
      \State \Return \Call{VoTE}{$F_d, \absvec{x}, pc_d$} = $Pass$
    \EndFunction
  \end{algorithmic}
\end{algorithm}
The algorithm defines two nested property checkers ($pc_d$ and $pc_i$), the
latter nested inside the former. Execution of the algorithm starts at line~22, 
where VoTE is invoked with the $d$-th tree ensemble, the given abstract input 
tuple $\absvec{x}$, and the outmost property checker $pc_d$ (defined on 
lines~2--18). Once invoked, VoTE will apply its abstraction-refinement approach, 
and invoke $pc_d$ with pairs $(\absvec{x}_d, \abs{y}_d)$ such that 
$\absvec{x}_d \sqsubseteq \absvec{x}$, and 
$\abs{y}_d = \abs{f}(\absvec{x}_d; F_d)$. 
Next, execution continues on line~12. Here, all tree ensembles except $F_d$ 
are enumerated, and VoTE is invoked again (line~13), now with each enumerated 
tree ensemble $F_i$, the input region $\absvec{x}_d$, and the second property
checker $pc_i$ (defined on lines~3--11). VoTE will apply its 
abstraction-refinement approach yet again, but now invoking $pc_i$ with 
pairs $(\absvec{x}_i, \abs{y}_i)$ such that 
$\absvec{x}_i \sqsubseteq \absvec{x}_d \sqsubseteq \absvec{x}$, 
and $\abs{y}_i = \abs{f}(\absvec{x}_i; F_i)$. Inside $pc_i$ (lines~3--11), 
we compare the abstract output of the $i$-th ensemble with the $d$-th (line~4). 
If the comparison is inconclusive, we return $Unsure$, if it is True, we 
return $Pass$, otherwise we return $Fail$. Back at line~14, we check the 
outcome of the most recent invocation to VoTE. If the outcome is $Fail$, there 
are some inputs captured by $\absvec{X}$ where the $i$-th tree ensemble 
predicts a larger value than the $d$-th ensemble, in which case the oracle
returns $False$. If the outcome is unsure, we instruct VoTE to further
refine $\absvec{X}_d$. Otherwise, we continue with the remaining tree 
ensembles.

The following Figure illustrates a sequence diagram, describing how the
different functions in the oracle interact with each other when all property
checkers return $Pass$.
\begin{figure}[ht]
  \center
   \includegraphics[scale=0.8]{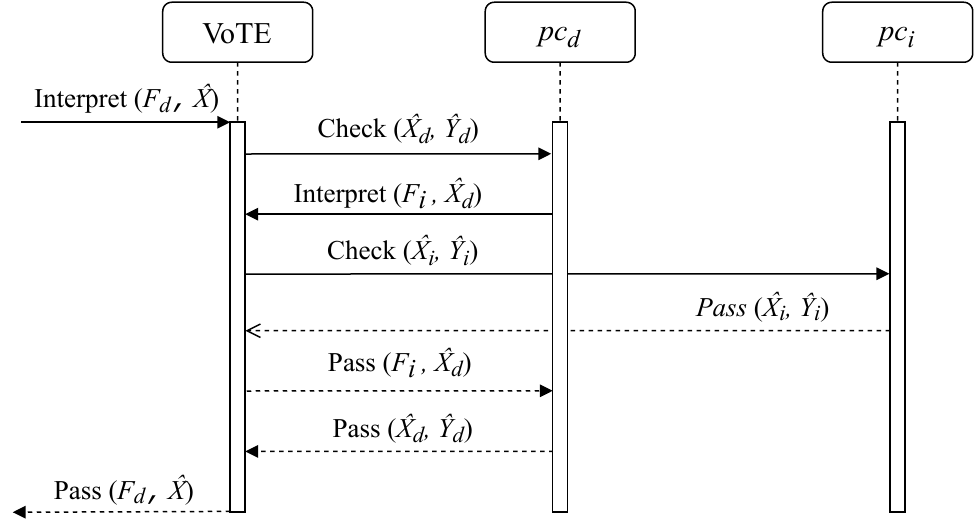}
   \caption{A sequence diagram exemplifying how functions are invoked (solid arrows)
            and returned from (dashed arrows) in Algorithm~\ref{algo:multi-oracle}
            when all property checkers return $Pass$. Text associated with arrows 
            describe an action taking place at a particular point in the sequence, 
            with parameters enclosed in parentheses.}
\end{figure}

\section{Other Algorithms for Computing Minimum Explanations}
\label{sec:minimum-alt}
In this section, we formalize two additional algorithms for computing
explanations that are minimum with respect to $g(I; W) = \sum_{i \in I}w_i$, 
where $I \subseteq \{1, \ldots, n\}$ and $W \in \R_{\ge 0}^n$. 
This first algorithm is based on a branch-and-bound approach realized in some
SMT solvers\cite{Dillig12}, and the second one is based on minimum
hitting sets\cite{Ignatiev15}, both of which were recently used to compute
minimum explanation in the context of neural networks\cite{La21}.
Out formalization of those two algorithms accepts the same inputs as m-MARCO, 
i.e., a tree ensemble $F$, the set of variables referenced by the ensemble $V_f$,
the inputs used in the prediction $(c_1, \ldots, c_n)$, the predicted label $d$,
and the weights for each feature $W$ used by the cost function.

\subsection*{Minimum Hitting Sets}
\label{sec:minimum-mhs}
Algorithm~\ref{algo:minimum-mhs} starts by defining the same front-facing 
oracle interface as used in m-MARCO, i.e., a function that accepts as input a 
set of variable indices $S$ to remove from a valid explanation (lines~2--8).
That function starts by initializing a tuple of abstract values 
$(\abs{x}_1, \ldots, \abs{x}_n)$ which capture all possible inputs to the 
tree ensemble (line~3). It then enumerates all variables referenced by the
tree ensemble, except those to exclude from the explanation, and tightens 
their abstractions to only include the value each enumerated variable had 
during the prediction (lines~4--6). Finally, the function invokes the 
underlying oracle (line~7).

Next, the algorithm initializes a collection of sets $M$ to hit with the 
empty set (line~9), and repeatedly computes a hitting set $I$ that is 
minimum with respect to the cost function $g$ (line~11), and also 
potentially a minimum explanation. To determine the validity of $I$ with 
the same oracle as used by m-MARCO, the algorithm computes the complement
of $I$ as $S = V_f \backslash I$, which becomes a seed with indices to 
delete from a valid explanation (line~12). If the oracle concludes that 
$S$ is valid, the algorithm terminates by returning $I$ which is a minimum
explanation. Otherwise, it shrinks the seed $S$ with indices to delete 
into a MUS (line~16), adds it to the collection of sets to hit (line~19),
and continues with the computation of another hitting set (line~11). This 
process is then repeated until a minimum hitting set that is also a valid
explanation is found.
\begin{algorithm}[ht]
  \caption{A minimum hitting set approach for computing an explanation that is 
           minimum with respect to the cost function $g$ for a tree ensemble
           $F$ and input sample $c_1, \ldots, c_n$, classified as label $d$.}
  \label{algo:minimum-mhs}

  \begin{algorithmic}[1]
    \Function{Minimum\_Explanation}{$F, V_f, c_1, \ldots, c_n, d, W=\{1\}^n$}
        \Function{Oracle}{$S$}
          \Comment{Define an oracle interface for MHS}
          \Let{$\abs{x}_1, \ldots, \abs{x}_n$}{$\top, \ldots, \top$}
          \Comment{Initialize all abstractions}
          \ForEach{$i \in V_f \backslash S$}
          \Comment{Compute MCS}
            \Let{$\abs{x}_i$}{$\alpha(\{c_i\})$}
            \Comment{Assign value to the $i$-th variable}
          \EndForEach
          \State \Return{\Call{Is\_Valid}{$F, (\abs{x}_1, \ldots, \abs{x}_n), d$}}
          \Comment{Query the oracle}
        \EndFunction

      \Let{$M$}{$\emptyset$}
      \Comment{Initialize sets to hit}
      \While{$True$}
        \Let{$I$}{\Call{MinimumHS}{$M, W$}}
        \Comment{Compute a minimum hitting set}
        \Let{$S$}{$V_f \backslash I$}
        \Comment{Compute a seed with indices to delete}
        \If{\Call{Oracle}{$S$}}
          \Comment{Check validity with the oracle}
          \State \Return $I$
          \Comment{$S$ is satisfiable, hence $I$ is a minimum explanation}
        \Else
          \Let{$S$}{\Call{Shrink}{$S$}}
          \Comment{$S$ is unsatisfiable, shrink it into a MUS}
          \Let{$M$}{$M \cup \big\{S\big\}$}
          \Comment{Add $S$ to the collection of sets to hit}
        \EndIf
      \EndWhile
    \EndFunction
  \end{algorithmic}
\end{algorithm}

\subsection*{Branch and Bound}
\label{sec:minimum-bb}

Algorithm~\ref{algo:minimum-bb} starts by initializing an explanation $I_{cur}$ 
with the indices of all variables referenced by the ensemble (line~1). It then 
initializes a tuple $(\abs{x}_1, \ldots, \abs{x}_n)$ that captures the entire 
concrete input domain (line~2). Next, a nested and recursive branch-and-bound 
procedure is defined (lines~4--16) that takes two input parameters: a set of 
indices $I$ of explanation variables, and a new index $i$ to be added to this 
set.

In this procedure, we start by adding the index $i$ to the set $I$ (line~5), 
and then bound the search for explanations to those that cost less than the 
current candidate $I_{cur}$ using the cost function $g$ (lines~6--8). 
If the cost $g(I)$ is greater than the cost $g(I_{cur})$, we simply return 
from the procedure. Otherwise, we refine the $i$-th abstraction to capture 
precisely the $i$-th concrete input value (line~9). We then query the oracle 
to see if all concrete values captured by the updated $(\abs{x}_1, \ldots, \abs{x}_n)$ 
map to the same label (line~10). 
If that is the case, we have found a cheaper explanation which we assign to
$I_{cur}$ (line~11). Otherwise, we permutate over the remaining referenced variable 
indices not in $I$ by invoking the recursive procedure (lines~13--15). 
Finally, the abstraction $\abs{x}_i$ is restored (line~17).

With the branch-and-bound procedure defined, we start the minimization of the
current candidate $I_{cur}$ by sequentially invoking the procedure with 
indices of variables referenced by the tree ensemble (lines~18--20). 
Finally, we return $I_{cur}$ (line~22).
\begin{algorithm}[ht]
  \caption{A branch and bound approach for computing an explanation that is 
           minimum with respect to the cost function $g$ for a tree ensemble
           $F$ and input sample $c_1, \ldots, c_n$, classified as label $d$.}
  \label{algo:minimum-bb}
  \begin{algorithmic}[1]
    \Function{Minimum\_Explanation}{$F, V_f, c_1, \ldots, c_n, d, W=\{1\}^n$}
      \Let{$I_{cur}$}{$V_f$}
      \Comment{Initialize current explanation to the most expensive}
      \Let{$\abs{x}_1, \ldots, \abs{x}_n$}{$\top, \ldots, \top$}
      \Comment{Relax all variables}

      \Procedure{Add\_Variable}{$I, i$}
        \Comment{Define recursive branch-and-bound procedure}
        \Let{$I$}{$I \cup \{i\}$}
        \Comment{Add index $i$ to set of indices $I$}
        \If{$g(I_{cur}; W) \le g(I; W)$}
        \Comment{Bound the search}
        \State \Return
        \EndIf
        \Let{$\abs{x}_i$}{$\alpha(\{c_i\})$}
        \Comment{Assign value to the $i$-th variable}
        \If{\Call{Is\_Valid}{$F, (\abs{x}_1, \ldots, \abs{x}_n), d$}}
          \Comment{Query the oracle}
          \Let{$I_{cur}$}{$I$}
          \Comment{Found a cheaper explanation}
        \Else
          \ForEach{$j \in V_f \backslash I$}
            \Comment{Branch on remaining indices}
              \State \Call{Add\_Variable}{$I, j$}
          \EndForEach
        \EndIf
        \Let{$\abs{x}_i$}{$\top$}
        \Comment{Relax the $i$-th variable again}
      \EndProcedure
      \ForEach{$i \in V_f$}
        \Comment{Start minimizing $I_{cur}$}
        \State \Call{Add\_Variable}{$\emptyset, i$}
      \EndForEach
      \State \Return $I_{cur}$
      \Comment{Return the set of indices in the explanation}
    \EndFunction
  \end{algorithmic}
\end{algorithm}

\section{Detailed Performance Results}
\label{sec:detailed-results}
In this section, we provide information on the runtime performance of algorithms
evaluated in Section~\ref{sec:comp-study}, but here for each individual model. 
In particular, Table~\ref{tbl:perf-enumerate} lists performance results for 
the standard MARCO approach (Algorithm~\ref{algo:marco}) which enumerates
all minimal explanations, and 
Tables~\ref{tbl:perf-minimum-marco}--\ref{tbl:perf-minimum-bb}
for computing a single minimum explanation using 
m-MARCO (Algorithm~\ref{algo:minimum-marco}),
MHS (Algorithm~\ref{algo:minimum-mhs}),
and BB (Algorithm~\ref{algo:minimum-bb}), respectively.

\begin{table}[ht]
  \centering
    \caption{Detailed performance results of Algorithm~\ref{algo:marco} (MARCO).}
  \begin{tabular}{|r|r|r|r|r|r|r|}
    \hline
    \multicolumn{1}{|r|}{\textbf{Name of}} &
    \multicolumn{2}{|r|}{\textbf{Input variables}} &
    \multicolumn{3}{c|}{\textbf{Runtime (s)}}  &
    \multicolumn{1}{c|}{\textbf{Number of}}   \\
    \cline{2-6} 
    \textbf{dataset} & \textbf{referenced} & \textbf{total} & 
    \textbf{min} & \textbf{avg} & \textbf{max} & 
     \textbf{timeouts} \\   
    \hline 
     ann-thyroid &        11 &    21 &    0.01 &     0.01 &     0.12  &      0 \\
    appendicitis &         7 &     7 &    0.00 &     0.00 &     0.01  &      0 \\
  biodegradation &        34 &    41 &    0.00 &   309.61 &  3,531.86  &     30 \\
         divorce &        14 &    54 &    0.00 &     0.02 &     0.04  &      0 \\
           ecoli &         6 &     7 &    0.00 &     0.02 &     0.05  &      0 \\
          glass2 &         8 &     9 &    0.00 &     0.00 &     0.01  &      0 \\
      ionosphere &        31 &    34 &    0.03 &   102.07 &  1,754.01  &      0 \\
       pendigits &        16 &    16 &    0.29 &    13.77 &    46.79  &      0 \\
       promoters &         1 &    58 &    0.00 &     0.00 &     0.00  &      0 \\
    segmentation &        17 &    19 &    0.01 &     0.70 &     5.64  &      0 \\
         shuttle &         9 &     9 &    0.00 &     0.02 &     0.07  &      0 \\
           sonar &        50 &    60 &    0.55 &  1,030.94 &  3,570.88  &     92 \\
        spambase &        49 &    57 &    0.05 &   735.89 &  3,130.13  &    141 \\
         texture &        37 &    40 &  147.60 &  1,902.62 &  3,544.87  &    135 \\
        threeOf9 &         3 &     9 &    0.00 &     0.00 &     0.00  &      0 \\
         twonorm &        20 &    20 &    0.00 &     8.53 &    79.86  &      0 \\
           vowel &        13 &    13 &    0.02 &     0.40 &     2.40  &      0 \\
            wdbc &        27 &    30 &    0.00 &    11.57 &    63.38  &      0 \\
wine-recognition &        12 &    13 &    0.00 &     0.02 &     0.09  &      0 \\
            wpbc &        32 &    33 &    0.06 &   227.72 &  2,683.32  &      1 \\
             zoo &        13 &    16 &    0.00 &     0.01 &     0.04  &      0 \\
    \hline
  \end{tabular}
  \label{tbl:perf-enumerate}
\end{table}

\begin{table}[ht]
  \centering
    \caption{Detailed performance results of Algorithm~\ref{algo:minimum-marco} (m-MARCO).}
  \begin{tabular}{|r|r|r|r|r|r|r|}
    \hline
    \multicolumn{1}{|r|}{\textbf{Name of}} &
    \multicolumn{2}{|r|}{\textbf{Input variables}} &
    \multicolumn{3}{c|}{\textbf{Runtime (s)}}  &
    \multicolumn{1}{c|}{\textbf{Number of}}   \\
    \cline{2-6} 
    \textbf{dataset} & \textbf{referenced} & \textbf{total} & 
    \textbf{min} & \textbf{avg} & \textbf{max} & 
     \textbf{timeouts} \\   
    \hline 
     ann-thyroid &        11 &    21 &    0.01 &     0.04 &     0.10  &      0 \\
    appendicitis &         7 &     7 &    0.00 &     0.00 &     0.00  &      0 \\
  biodegradation &        34 &    41 &    0.02 &    27.97 &   352.61  &      0 \\
         divorce &        14 &    54 &    0.00 &     0.01 &     0.01  &      0 \\
           ecoli &         6 &     7 &    0.01 &     0.02 &     0.04  &      0 \\
          glass2 &         8 &     9 &    0.00 &     0.00 &     0.01  &      0 \\
      ionosphere &        31 &    34 &    0.04 &     2.12 &    13.03  &      0 \\
       pendigits &        16 &    16 &    0.21 &     2.12 &     8.01  &      0 \\
       promoters &         1 &    58 &    0.00 &     0.00 &     0.00  &      0 \\
    segmentation &        17 &    19 &    0.01 &     0.14 &     0.61  &      0 \\
         shuttle &         9 &     9 &    0.00 &     0.02 &     0.06  &      0 \\
           sonar &        50 &    60 &    0.15 &    31.72 &   277.70  &      0 \\
        spambase &        49 &    57 &    0.12 &   227.75 &  2,770.22  &      1 \\
         texture &        37 &    40 &    1.91 &    31.93 &   192.84  &      0 \\
        threeOf9 &         3 &     9 &    0.00 &     0.00 &     0.00  &      0 \\
         twonorm &        20 &    20 &    0.00 &     0.84 &     5.96  &      0 \\
           vowel &        13 &    13 &    0.04 &     0.19 &     0.61  &      0 \\
            wdbc &        27 &    30 &    0.02 &     0.29 &     1.10  &      0 \\
wine-recognition &        12 &    13 &    0.00 &     0.02 &     0.03  &      0 \\
            wpbc &        32 &    33 &    0.05 &     1.31 &    15.01  &      0 \\
             zoo &        13 &    16 &    0.00 &     0.01 &     0.04  &      0 \\
    \hline
  \end{tabular}
  \label{tbl:perf-minimum-marco}
\end{table}

\begin{table}[ht]
  \centering
    \caption{Detailed performance results of Algorithm~\ref{algo:minimum-mhs} (MHS).}
  \begin{tabular}{|r|r|r|r|r|r|r|}
    \hline
    \multicolumn{1}{|r|}{\textbf{Name of}} &
    \multicolumn{2}{|r|}{\textbf{Input variables}} &
    \multicolumn{3}{c|}{\textbf{Runtime (s)}}  &
    \multicolumn{1}{c|}{\textbf{Number of}}   \\
    \cline{2-6} 
    \textbf{dataset} & \textbf{referenced} & \textbf{total} & 
    \textbf{min} & \textbf{avg} & \textbf{max} & 
     \textbf{timeouts} \\   
    \hline 
     ann-thyroid &        11 &    21 &    0.03 &     0.04 &     0.13  &      0 \\
    appendicitis &         7 &     7 &    0.00 &     0.00 &     0.01  &      0 \\
  biodegradation &        34 &    41 &    0.05 &    66.49 &  1,141.11  &      0 \\
         divorce &        14 &    54 &    0.00 &     0.01 &     0.02  &      0 \\
           ecoli &         6 &     7 &    0.01 &     0.03 &     0.05  &      0 \\
          glass2 &         8 &     9 &    0.00 &     0.01 &     0.01  &      0 \\
      ionosphere &        31 &    34 &    0.06 &     6.62 &    40.42  &      0 \\
       pendigits &        16 &    16 &    0.48 &     3.48 &    15.31  &      0 \\
       promoters &         1 &    58 &    0.00 &     0.00 &     0.00  &      0 \\
    segmentation &        17 &    19 &    0.01 &     0.17 &     0.91  &      0 \\
         shuttle &         9 &     9 &    0.01 &     0.02 &     0.08  &      0 \\
           sonar &        50 &    60 &    0.32 &   101.19 &   860.65  &      0 \\
        spambase &        49 &    57 &    0.13 &   521.09 &  3,170.05  &      8 \\
         texture &        37 &    40 &    3.56 &    86.84 &   918.36  &      0 \\
        threeOf9 &         3 &     9 &    0.00 &     0.00 &     0.00  &      0 \\
         twonorm &        20 &    20 &    0.02 &     2.21 &    12.38  &      0 \\
           vowel &        13 &    13 &    0.10 &     0.35 &     0.91  &      0 \\
            wdbc &        27 &    30 &    0.04 &     0.63 &     2.35  &      0 \\
wine-recognition &        12 &    13 &    0.00 &     0.02 &     0.06  &      0 \\
            wpbc &        32 &    33 &    0.09 &     3.65 &    40.78  &      0 \\
             zoo &        13 &    16 &    0.01 &     0.03 &     0.07  &      0 \\
    \hline
  \end{tabular}

  \label{tbl:perf-minimum-mhs}
\end{table}

\begin{table}[ht]
  \centering
    \caption{Detailed performance results of Algorithm~\ref{algo:minimum-bb} (BB).}
  \begin{tabular}{|r|r|r|r|r|r|r|}
    \hline
    \multicolumn{1}{|r|}{\textbf{Name of}} &
    \multicolumn{2}{|r|}{\textbf{Input variables}} &
    \multicolumn{3}{c|}{\textbf{Runtime (s)}}  &
    \multicolumn{1}{c|}{\textbf{Number of}}   \\
    \cline{2-6} 
    \textbf{dataset} & \textbf{referenced} & \textbf{total} & 
    \textbf{min} & \textbf{avg} & \textbf{max} & 
     \textbf{timeouts} \\   
    \hline 
     ann-thyroid &        11 &    21 &    0.07 &     6.71 &   229.11  &      0 \\
    appendicitis &         7 &     7 &    0.00 &     0.01 &     0.19  &      0 \\
  biodegradation &        34 &    41 &      -- &       -- &     --    &    200 \\
         divorce &        14 &    54 &    0.00 &     0.06 &     0.47  &      0 \\
           ecoli &         6 &     7 &    0.02 &     0.11 &     0.60  &      0 \\
          glass2 &         8 &     9 &    0.00 &     0.06 &     0.48  &      0 \\
      ionosphere &        31 &    34 &    0.07 &   260.19 &  3,475.98  &    153 \\
       pendigits &        16 &    16 &  3,080.18 &  3,080.18 &  3,080.18  &    199 \\
       promoters &         1 &    58 &    0.00 &     0.00 &     0.00  &      0 \\
    segmentation &        17 &    19 &    0.04 &   214.99 &  2,140.51  &     53 \\
         shuttle &         9 &     9 &    0.02 &     0.22 &     0.98  &      0 \\
           sonar &        50 &    60 &      -- &       -- &     --    &    200 \\
        spambase &        49 &    57 &      -- &       -- &     --    &    200 \\
         texture &        37 &    40 &      -- &       -- &     --    &    200 \\
        threeOf9 &         3 &     9 &    0.00 &     0.00 &     0.00  &      0 \\
         twonorm &        20 &    20 &      -- &       -- &     --    &    200 \\
           vowel &        13 &    13 &   17.93 &   781.71 &  2,472.43  &     95 \\
            wdbc &        27 &    30 &   38.17 &   809.89 &  3,551.16  &     91 \\
wine-recognition &        12 &    13 &    0.01 &     0.27 &     4.36  &      0 \\
            wpbc &        32 &    33 &   75.08 &  2,241.61 &  3,561.59  &    136 \\
             zoo &        13 &    16 &    0.02 &   246.66 &  2,001.73  &      0 \\
    \hline
  \end{tabular}
  \label{tbl:perf-minimum-bb}
\end{table}

\end{document}